\documentclass{article} 
\usepackage{iclr2024_conference,times}

\usepackage{amsmath,amsfonts,bm}

\def\eqref#1{equation~\ref{#1}}

\def\1{\bm{1}}

\DeclareMathAlphabet{\mathsfit}{\encodingdefault}{\sfdefault}{m}{sl}
\SetMathAlphabet{\mathsfit}{bold}{\encodingdefault}{\sfdefault}{bx}{n}

\usepackage[most]{tcolorbox}
\usepackage{url}
\usepackage{graphicx}
\graphicspath{{image/}}
\usepackage{enumitem}
\usepackage{ulem}
\usepackage{framed}
\usepackage{color}
\usepackage{xcolor}
\usepackage{subfigure}
\usepackage{caption,subcaption}
\usepackage{listings}
\usepackage[thicklines]{cancel}
 
\usepackage{wrapfig}
\usepackage{setspace}
\usepackage{mathrsfs}
\usepackage{multirow}
\usepackage{amssymb}
\usepackage{threeparttable}
\usepackage{ulem} 
\definecolor{citecolor}{HTML}{0071BC}
\definecolor{linkcolor}{HTML}{ED1C24}
\definecolor{shadecolor}{rgb}{0.92,0.92,0.92}
\usepackage[pagebackref=false,breaklinks=true,colorlinks,citecolor=citecolor,linkcolor=linkcolor,bookmarks=false]{hyperref}

\definecolor{blue}{rgb}{0, 0, 1} 
\usepackage{listings}
\definecolor{codegreen}{rgb}{0,0.6,0}
\definecolor{codepurple}{rgb}{0.58,0,0.82}
\lstdefinestyle{pythonstyle}{
    commentstyle=\color{black},
    keywordstyle=\color{blue},
    numberstyle=\tiny\color{codegray},
    stringstyle=\color{codepurple},
    basicstyle=\small\ttfamily\footnotesize\bfseries,
    breakatwhitespace=false,
    breaklines=true,                 
    captionpos=b,                    
    keepspaces=true,                  
    showspaces=false,                
    showstringspaces=false,
    showtabs=false,                  
    tabsize=2
}
\title{InstructDET: Diversifying Referring Object Detection with Generalized Instructions}

\iclrfinalcopy

\author{
\textbf{Ronghao Dang}$^{1\ast}$\quad Jiangyan Feng$^2$\thanks{R. Dang and J. Feng contribute equally. $^\star$Y. Song is the corresponding author. This work is done when R. Dang is an intern at Sensetime. The code is available at \url{https://github.com/jyFengGoGo/InstructDet}. } \quad
\textbf{Haodong Zhang}$^2$ \quad
\textbf{Chongjian GE}$^3$ \quad 
\textbf{Lin Song}$^4$ \\
\textbf{Lijun Gong}$^2$ \quad 
\textbf{Chengju Liu}$^1$ \quad 
\textbf{Qijun Chen}$^1$ \quad 
\textbf{Feng Zhu}$^2$ \quad 
\textbf{Rui Zhao}$^2$ \quad 
\textbf{Yibing Song}$^{5\star}$\\
$^1$Tongji University \quad
$^2$SenseTime Research \quad
$^3$The University of Hong Kong\\
$^4$Tencent AI Lab\quad
$^5$Alibaba DAMO Academy\\
{\tt\scriptsize 
dangronghao@tongji.edu.cn\quad 
fengjiangyan@sensetime.com \quad
yibingsong.cv@gmail.com
}
}

\begin{document}

\maketitle
\vspace{-10mm}
\begin{figure}[h]
    \centering
    \includegraphics[width=0.8\textwidth]{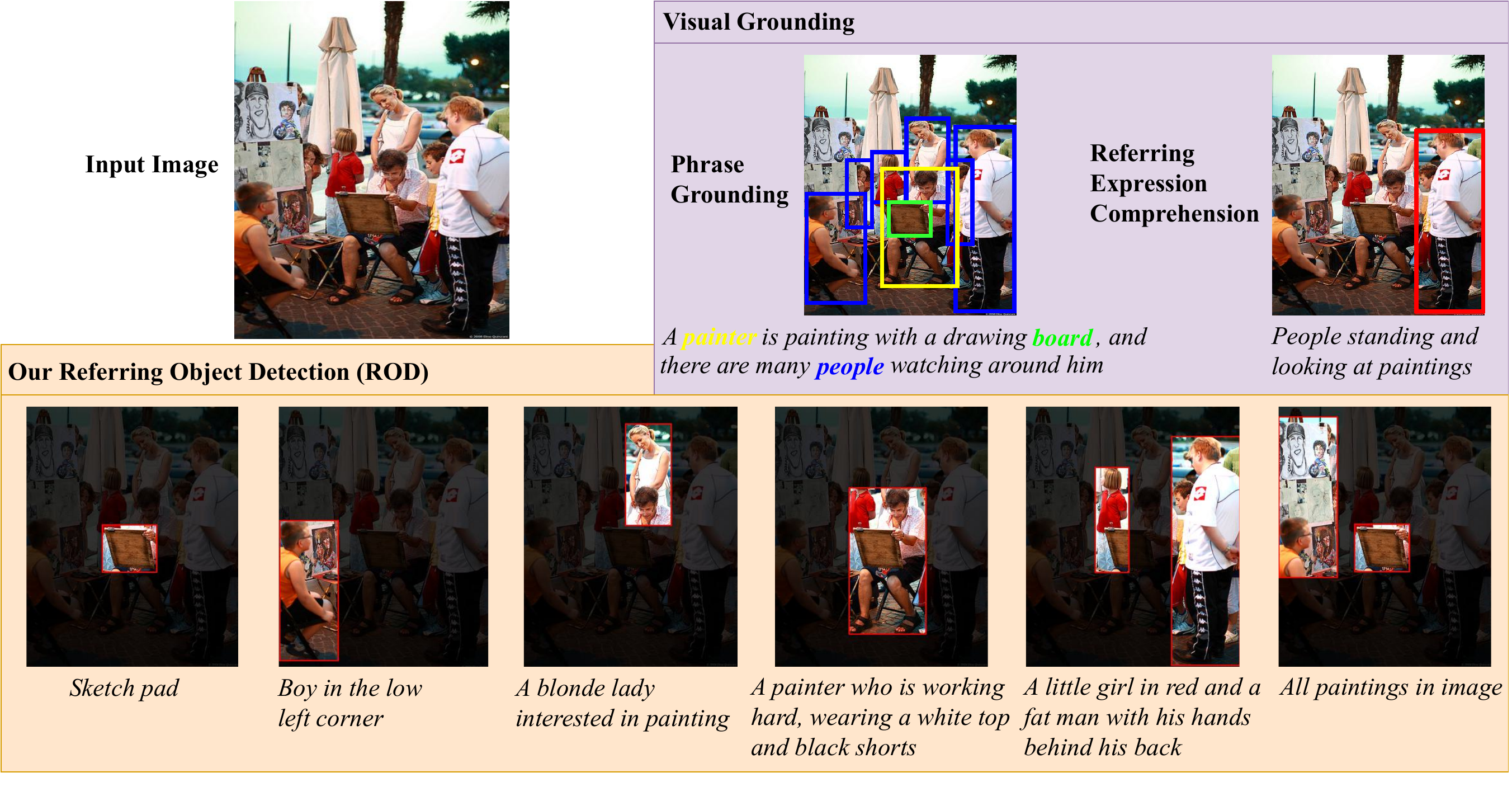}
    \vspace{-5mm}
    \caption{Our ROD aims to execute diversified user detection instructions compared to visual grounding. For images with object bbxs, we use foundation models to produce human-like object detection instructions. By training a conventional ROD model with incorporating tremendous instructions, we largely push ROD towards practical usage from a data-centric perspective.}
    \label{fig:teaser}
\end{figure}

\begin{abstract}
We propose InstructDET, a data-centric method for referring object detection (ROD) that localizes target objects based on user instructions. While deriving from referring expressions (REC), the instructions we leverage are greatly diversified to encompass common user intentions related to object detection. For one image, we produce tremendous instructions that refer to every single object and different combinations of multiple objects. Each instruction and its corresponding object bounding boxes (bbxs) constitute one training data pair. In order to encompass common detection expressions, we involve emerging vision-language model (VLM) and large language model (LLM) to generate instructions guided by text prompts and object bbxs, as the generalizations of foundation models are effective to produce human-like expressions (e.g., describing object property, category, and relationship). We name our constructed dataset as InDET. It contains images, bbxs and generalized instructions that are from foundation models. Our InDET is developed from existing REC datasets and object detection datasets, with the expanding potential that any image with object bbxs can be incorporated through our InstructDET method. By using our InDET dataset, we show that a conventional ROD model surpasses existing methods on both standard REC datasets and our InDET test set. InstructDET, our data-centric method with automatic data expansion by leveraging foundation models, directs a promising field that ROD can be greatly diversified to execute common object detection instructions.  
\end{abstract}

\section{Introduction}

Referring object detection (ROD) aims to detect target objects according to language reference that represents user intentions. ROD is closely related to visual grounding where there are phrase grounding~\citep{multi-level_IPG,Li_2022_CVPR,gvlm_2023_iclr} and referring expression comprehension~\citep{vlbert_iclr20,seqtr}. As shown in Fig.~\ref{fig:teaser}, phrase grounding detects all objects mentioned in one sentence, while referring expression comprehension (REC) only detects one single object that the text refers to. As such, the language reference in REC shall be discriminative and specifically relates to one object without ambiguity.

Currently, visual grounding develops at an initial stage and leaves a gap for practical usage. The phrase grounding does not differentiate which object ought to be detected via language description, while REC only targets for one object with single text reference. In the current REC datasets, each image contains few expressions (e.g., 1 or 2 phrases). These expressions are insufficient to represent user intentions. In an image where there are several objects, users may want to detect each single object by using different descriptions (e.g., object color, shape, or location), or detect multiple objects in different combinations (e.g., similar properties or relationships). These diverse expressions are not conveyed within current REC datasets, leaving the gap for existing methods to practically fulfill user intentions for visual grounding. Moreover, the manual collection of these expressions are cumbersome, and subject bias prevents an effective coverage of common user intentions when perceiving each image. Therefore, the practical user expressions are not well fulfilled when they expect to detect various objects in one image.

In this work\footnote{We do not differentiate ``instruction'' and ``expression'' in this paper, as both of them represent user intentions. For presentation clarity, in our InstructDET pipeline we refer expressions that are generated by foundation models, and we further refine expressions to instructions for InDET inclusion. As we only focus on ROD, we can formalize our instruction by simply adding the word `detect' beforehand.}, we aim to push visual grounding toward practical usage from a data-centric perspective. Instead of developing REC models to generalize based on current data, we set up referring object detection (ROD) scenario to automatically diversify user expressions. Inspired by the generalization of foundation models that execute common user instructions based on the image and text inputs, our InstructDET borrows their capabilities to produce human-like instructions that encompass user intentions related to object detection. The generalized instructions produced by the foundation models can be regarded as an expansion of existing user expressions in REC. We produce instructions that describe single object from two pipelines. In the first pipeline (i.e., global prompt), we convert an image into an elaborate text description via LLaVA~\citep{LLAVA}. The text description, together with object bbxs coordinates, are sent to the LLaMA~\citep{LLaMA} for instruction generation in global prompt. During generation, we manually write 3 in-context examples and leverage the in-context learning~\citep{icl} ability of LLaMA to describe the content related to each object following the format of our examples. 

In the second pipeline (i.e., local prompt), we send the image and text prompts into LLaVA. The objects in the image are marked with bbxs and the text prompts require LLaVA to describe the object content. We initialize LLaVA with miniGPT4 weights and find it tends to produce lengthy and global descriptions. So we perform a partial finetuning on LLaVA by using REC data to let it focus on local objects. Through these two pipelines, we observe that instructions generated from global prompt pipeline focus more on the object relationship, while instructions generated from local prompt pipeline focus more on rich visual details and advanced logic reasoning. Naturally, we combine instructions from these two pipelines to formulate expressions for single referred object. During instruction generation, the uncontrolled model hallucination~\citep{diffusion-lm} brings incorrect or irrelevant instructions. We propose to use visual-textual verification via CLIP~\citep{clip} for effective instruction filtering.

The generalization and reasoning of foundation models~\citep{llm-generalization,llm-reasoning} provide sufficient instructions encompassing user intentions for single object description. When describing multiple objects, we divide descriptions into two parts. The first part is to independently describe each single object followed by concatenation, and the second part is to summarize commonalities of multiple objects. The commonality summarization requires unifying similar or related objectives by a higher-level language abstraction that describes their similarities and relationships. We collect the combinations of different objects via semantic clustering, then utilize LLM to generate commonality summarizations for each combination.

We automatically collect instructions targeting for single or multiple objects in images and construct our InDET dataset. Sec.~\ref{sec:analysis} shows an in-depth analysis of our dataset where we establish a guideline to organize these instructions from 6 aspects. Compared to existing REC datasets where the instructions only reside in sub-parts of our groups, our InDET is more comprehensive to incorporate user intentions of object detection. 
Fig.~\ref{fig:teaser} shows an intuitive example of the generalized expressions produced by foundation models. By using our InDET dataset, we train a conventional ROD model and find it surpasses existing VG models on standard benchmarks and our InDET test set. Moreover, we also validate that our model 
has learned to effectively understand the meaning of instructions rather than only recognize key words, which is because of the tremendously expressive instructions incorporated for our model training. Our InstructDET method can automatically expand training data by using in-the-wild images with object bbxs, which improves our model generalizations towards practical usage. In addition, our model can already serve as the detection module of the neural-symbolic visual compositional task solution given arbitrary language instructions beyond object detection (e.g., Visual ChatGPT~\citep{visualchatgpt}, VISPROG~\citep{VISPROG_2023_CVPR}). 

\section{Related Works}
\vspace{-2mm}

{\bf Visual Grounding}.
Studies on visual grounding~\citep{mdetr,groundobj_2021_iclr,transVG_iccv21,Su_2023_CVPR} can be mainly categorized as phrase grounding~\citep{npas_iclr22,PGTP} and REC~\citep{comatt_iclr18,referTrans_nips21}. Phrase grounding detects all objects mentioned in the text while REC localizes one object that the text referred to. In~\citep{glipv2_nips22,gdino_arxiv23}, the objects mentioned in the text are verified to each visual object proposal one-by-one. These methods require a clear and specific object referring in the text. On the other hand, methods~\citep{seqtr,UNINEXT} based on DETR~\citep{DETR} can accept abstract and summarized descriptions such as ``red objects'' and ``all objects''. Our ROD model follows DETR-based design to enrich interpretation of various instructions. Note that our model is learned via InDET dataset where instructions are produced based on preset object bbxs. 


{\bf Referring Expression Datasets}.
The REC datasets are usually constructed via manual annotation on the images. A two-player game is utilized in~\citep{referitgame} where the text descriptions are concise due to limited relevant visual contents. The RefCOCO, RefCOCO+ \citep{refcoco}, and RefCOCOg~\citep{refcocog} employ MSCOCO~\citep{mscoco} images for manual expression production. The expression flexibility and diversity of these datasets are limited to encompass common detection intentions. Recent datasets~\citep{krishna2017visual,openimg-ijcv20,goldg-nips21} focuses on data scalability rather than auto generation. 
Cops-Ref \citep{cops-ref} leverages scene graph as reasoning groundwork, thus forming a tree structure to generate expressions with varying compositionality. Different from these methods based on template guided expression generation, our InstructDET relies on foundation models to produce well generalized and human-like instructions.


{\bf Data Generation via Foundation Models}.
The InstructGPT~\citep{instructgpt} and GPT4~\citep{gpt-4} have shown generalization and reasoning abilities for data generation. LLaVA~\citep{LLAVA} first uses GPT4 for multi-modal data generation following instructions. 
Otter~\citep{otter} performs multi-modality in-context instruction tuning by levering multiple images, questions, and answers. Currently, these models focus on global image and language understanding, with less focus on local object analysis. Moreover, these multi-modality models, although processing multi-modality data, still outputs single-modality text description. There is a gap for these foundation models to function in the computer vision scenarios, especially visual recognition. In comparison, our InstructDET uses foundation models to benefit ROD model training, which contributes directly to improve object detection performance.


\begin{figure}[t]
    \centering
    \includegraphics[width=\textwidth]{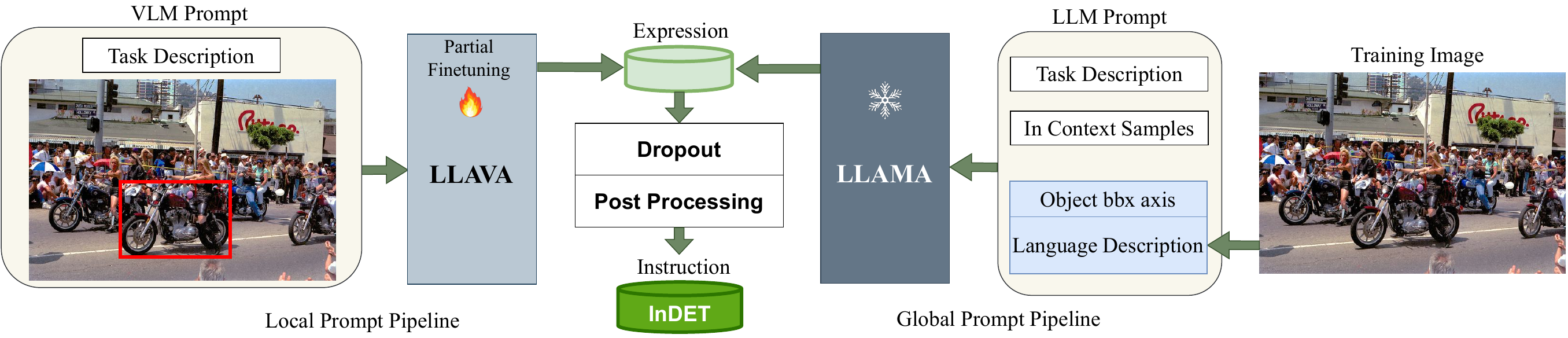}
    \vspace{-6mm}
    \caption{An overview of our InstructDET. We use two pipelines to produce detection expressions via foundation models. In the global prompt pipeline, we use LLaVA to describe an image via text, and combine this text with other text prompts for LLaMA input. In the local prompt pipeline, we use the same image with object bbxs and text prompts as multi modality input for LLaVA. The produced expressions are further refined to instructions and incorporated into our InDET dataset.}
    \label{fig:overview}
    \vspace{-10pt}
\end{figure}

\section{InstructDET}
Fig.~\ref{fig:overview} shows an overview of our InstructDET method for data construction. Given an input image with object bbxs, we use two pipelines to produce detection expressions from foundation models. The expressions are further refined to instructions and incorporated into our InDET dataset. 

\subsection{Global Prompt Pipeline}
The large language model (LLM) has shown surprising generalizations to well execute common user instructions. We use LLM to simulate user intentions when perceiving objects in an image. Our global prompt pipeline produces a text prompt for the LLaMA\footnote{In this paper, we use a variant of LLaMA (i.e., Vicuna 13B) that has gone through instruction tuning. Besides, we use a variant of LLaVA, which is a multi-modal paradigm that maps visual features into token embeddings with further alignment to the text domain.} model. This prompt consists of several contents including global image description, object bbx coordinates, in-context samples, and task description. Without tuning LLaMA, we obtain instructions that describe objects in an image. A detailed example is shown in Sec.~\ref{sec:supp-single-modality} for an intuitive illustration of how we leverage foundation models to produce expressions. We elucidate the key steps during this process as follows: 

Given an image with object bbxs, we first obtain global image description in text form. If this image already contains dense captions (e.g., from Flicker30K), we directly load these captions. 
Alternatively, we leverage LLaVA to generate the global image description.
The text prompt we use for LLaVA contains our language guidance to emphasize that specific interested object categories shall be mentioned. As such, LLaVA will describe each labeled object in its output. As for object bbx content, if the image is from REC dataset, we use referring expression as the object content. Otherwise, we simply use the category name.

When designing the task description prompt, we expect LLaMA to produce diverse expressions that contain different properties of single object as much as possible. We manually list the attributes from the user perspective, including the object type, color, function, motions, etc. Besides, we include the object attributes of its relationship with other objects in image, such as object interactions, object relative positions, etc. When using these aforementioned prompts for LLaMA input, we find that the output text varies significantly and might be irrelevant to the target objects. Inspired by the in-context learning ability of foundation models, we manually design in-context samples to regularize the output content and format. The output results will thus resemble our in-context examples but with our expected diversified object descriptions.

\subsection{Local Prompt Pipeline}
The global prompt pipeline produces expressions according to text prompts. Naturally, we can feed both image and text prompt to the multi-modality foundation model for object description. Given an image, we mark the object with bbx rectangle, and send this image to LLaVA, together with the text prompt that requires LLaVA to describe the object according to the bbx. Here, the bbx serves as a visual highlight for LLaVA to comprehend the target object that we expect to describe. 

Our LLaVA model is initialized with miniGPT4 weights. When we send these multi-modality inputs to LLaVA, we observe that LLaVA produces detailed and dense descriptions for the image, rather than expressions of the specific target object. We analyze that the vision-language alignment module in LLaVA is the Q-Former~\citep{blip-2}, which transforms one image into only 32 visual tokens without concentrating on the local objects. Meanwhile, LLaVA itself tends to produce lengthy and dense descriptions. In order to generate instructions suitable for ROD, we finetune a part of LLaVA by using existing REC datasets. Specifically, we only update a linear layer that transforms visual tokens to the text embedding space during training. The linear layer is learned to attend local objects with concise expressions. After finetuning, we observe the LLaVA output becomes informative and closely related to the target object. Detailed examples of generating expressions in local prompt pipeline are shown in Sec.~\ref{sec:supp-multi-modality}, and the detailed analysis on how finetuning improves LLaVA output is provided in Sec.~\ref{sec:supp-ablation}.

\subsection{Expression Filter}

\begin{wrapfigure}{R}{0.5\textwidth}
\vspace{-20pt}
\begin{center}
    \includegraphics[width=0.5\textwidth]{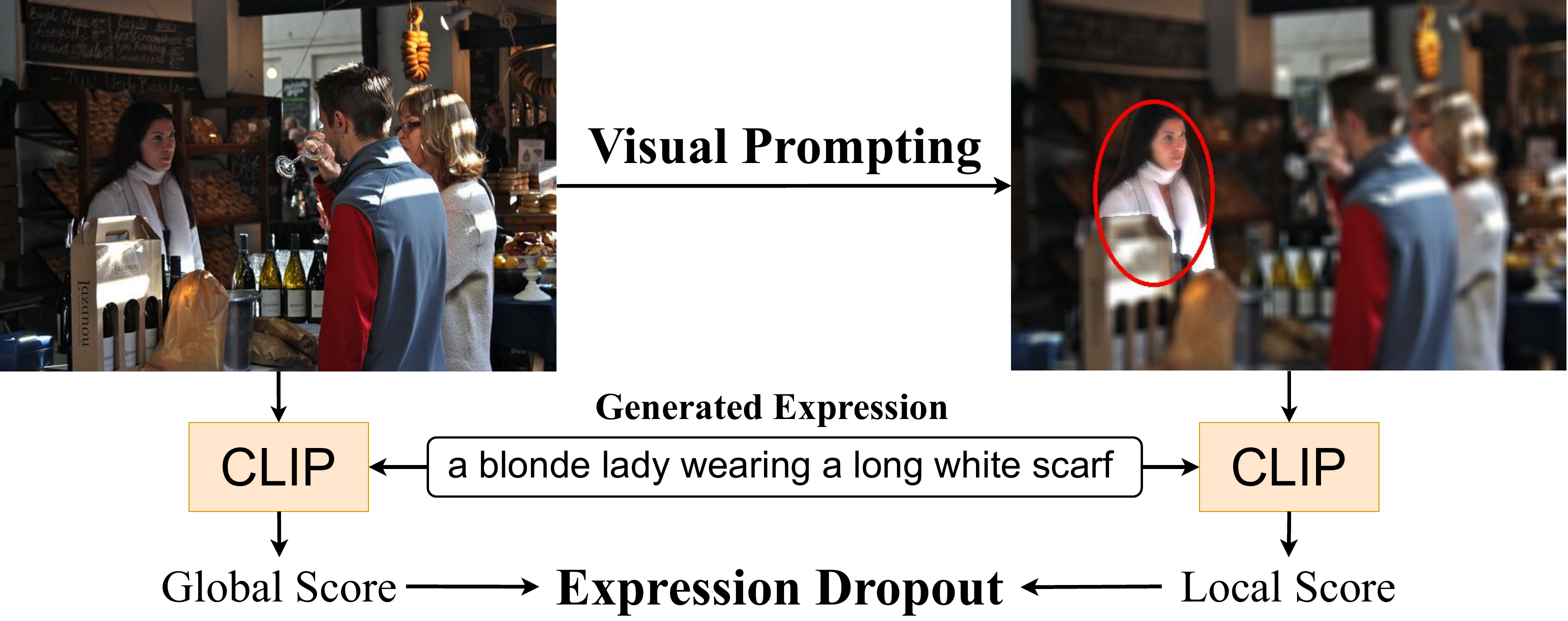}
    \vspace{-6mm}
    \caption{Expression filtering by image visual prompting and visual-textual matching via CLIP.}
    \label{fig:dropout}
\end{center}
\vspace{-10pt}
\end{wrapfigure}

In global and local pipelines, we have regularized the output of foundation model from several aspects including text prompt specification, in-context learning, and model finetuning. In practice, we still observe the model hallucination phenomena that the model sometimes generate expressions describing objects not even exist in the image. Moreover, the expressions from the local prompt pipeline sometimes describe the whole image rather than local objects. This is due to the miniGPT4 initialization of LLaVA, which utilizes dense captions for instruction tuning. The tendency to generate global image description is mitigated via our model finetuning to focus on local object, but not completely disappeared. To further improve the expression quality, we introduce visual and language matching via CLIP~\citep{clip} to filter out inappropriate expressions. Fig.~\ref{fig:dropout} shows an overview. It contains image visual prompting and visual-textual matching. 

\paragraph{Visual Prompting}
We study visual language pretraining (VLP)~\citep{FGVP,red_circle} where visual prompting is developed for images. We observe that in zero-shot REC, coupling VLP with visual prompts enables robust pairing of local image region and corresponding text description. In the pairing process, the design of visual prompting heavily influences the visual-textual matching results. 
Specifically, we employ the superposition of a red ellipse and the target Gaussian blur reversion as visual prompts. A detailed pipeline illustrating visual prompting is in Sec.~\ref{sec:supp-vprompt}.


{\bf Visual-Textual Matching}. 
We use images with visual prompting that emphasizes target objects to verify the corresponding text descriptions via a frozen CLIP model. While local object contents are well aligned with target referring expressions, we observe that expressions describing the whole image are not eliminated by CLIP. We analyze that CLIP is originally trained to focus on the correlations between global image features and global textual semantics. The global visual-textual matching makes CLIP model to prefer global image description accordingly. To remove this effect, we establish a referring measurement from both local and global perspectives. For the image in Fig.~\ref{fig:dropout}, we compute a global score $S_g$ and a local prompt score $S_l$. The magnitude of referring expression can be measured via our local enhancement score $S_e=S_l-S_g$. Our final expression evaluation score can be computed as:
\begin{equation}\label{eq:dropout}
S_f = \alpha_{1} S_e + \alpha_{2} S_l = (\alpha_{1} + \alpha_{2})S_l - \alpha_{1} S_g = S_l - \alpha_{1}S_g
\end{equation}
where $\alpha_1$ and $\alpha_2$ are scalars balancing the contributions of $S_g$ and $S_l$ with $\alpha_1+\alpha_2=1$. So $\alpha_{1} \in [0, 1]$ adjusts the final score towards local content referring or global semantics. Note that we introduce $S_e$ to measure difference between local and global scores. If the expression is more related to the target object, $S_e$ becomes higher after visual prompting for object highlights. After computing $S_f$, we set a dynamic threshold to filter out expressions. This is because $S_f$ is based on CLIP's preference that a small target object with well matched expression achieves a lower score than a large object with mismatched expression. Therefore, we use provided expression (for images from REC) or category name (for images out of REC) to compute a final score, and discard generated instructions whose $S_f$ is lower than this score.

\begin{wrapfigure}{R}{0.5\textwidth}
\vspace{-20pt}
\begin{center}
    \includegraphics[width=0.5\textwidth]{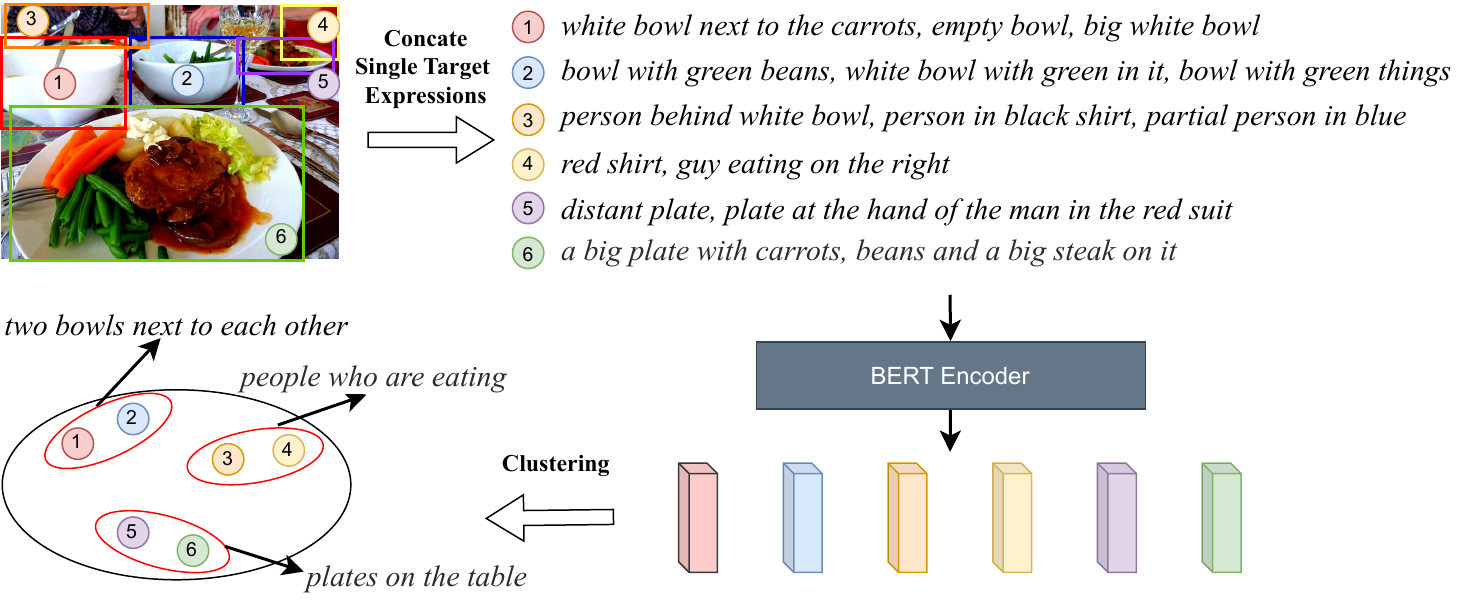}
    \caption{Mining commonalities among multi-objects via expression concatenation and text semantic clustering, followed by LLaMA descriptions on each cluster center.}
    \label{fig:multi_object}
\end{center}
\vspace{-10pt}
\end{wrapfigure}

\subsection{Multi-Objects Expression Generation}
\label{multi-target expression}
Our expression generation pipeline illustrated above targets for each object independently. In practice, users may refer to multiple objects in one image. We study common user expressions for multi-objects, and conclude them into two aspects. The first one contains splicing expressions that combine different single object expressions with `and' or comma. In this case, the objects mentioned in the expression are not related to each other. The second aspect contains generalization expressions that summarize the common properties of multiple objects (e.g., color, category, or location) to produce an abstract and conceptual description. It resembles mining similarities between multiple objects and thus is not straightforward to conclude. Therefore, we need to discover object combinations where similar properties may exist, and then summarize the commonalities among them to constitute the summarized expressions. 

Our process to produce summarized expression is shown in Fig.~\ref{fig:multi_object}. For each object, we first concatenate all its related expressions with commas. Through this concatenation, we can obtain this object expression from different perspectives (i.e., different properties). Then, we use the text encoder in BERT~\citep{bert} to map this concatenated expression to a semantic embedding space. As such, we obtain embeddings of concatenated expressions from all objects. Then, we cluster these embeddings into indeterminate number of clusters by using DBSCAN~\citep{DBSCAN} method. We use LLaMA to generate text for clusters with multiple objects. The details of using LLaMA to mine object commonalities are in Sec.~\ref{sec:supp-multiobj}. The generated text indicates the summarized expression we aim to produce for multiple objects.

{\bf Post Processing}. 
After generating expressions for single and multiple objects, we verify and remove the repeated expressions that pertain to the same object. Then, we utilize LLaMA to further diversify these generated expressions while preserving their original intents, i.e., we use LLaMA to do synonymous rewriting of generated expressions. The prompt we use for synonymous rewriting is provided in Sec.~\ref{sec:supp-syn-rewrite}. We observe that for different objects in one image, the expression for one object may be similar to that of others. These expressions are ambiguous since we can not refer to a unique object based on their referring. Nevertheless, we transfer these expressions to refer to multi-objects since they express a set of objects in one image. This transfer further augments multi-object referring expressions. Finally, we collect these remaining expressions after post processing as instructions. Together with corresponding object bbxs and images, we construct our InDET dataset by incorporating diversified object detection instructions encompassing user intentions.

\section{Dataset Analysis}\label{sec:analysis}

Our InDET dataset contains images from MSCOCO~\citep{mscoco}, Flicker~\citep{flickr30k}, and Objects365~\citep{objects365}. There are 120.6K images with 908.4K referring object sets in total. Together with original expressions, there are 3.6M instructions in total, making InDET the largest real-world REC dataset at present. The average instruction length is 6.2 words and the vocabulary size is 63k words, which surpasses existing automatically annotated datasets in terms of instruction quantity, richness, and vocabulary breadth. We split the images into training, validation, and testing sets, with the corresponding instruction amount of 3139K, 240K, and 247K, respectively. In the following, we first propose a guideline that represent common user intentions and divides existing instructions into 6 groups. Then, we analyze all the instructions in InDET according to this guideline to show how our InDET advances REC scenario compared to existing datasets.

\begin{table}[t]
    \centering
    \vspace{-10mm}
    \caption{Instruction guideline and samples. Our guideline contains 6 aspects and covers common user intentions. These aspects are built upon object category, attribute, and relations with different emphasis levels. We use $\star$ and $\star \star$ to indicate the description complexity of different aspects.}   
    \includegraphics[width=0.8\textwidth]{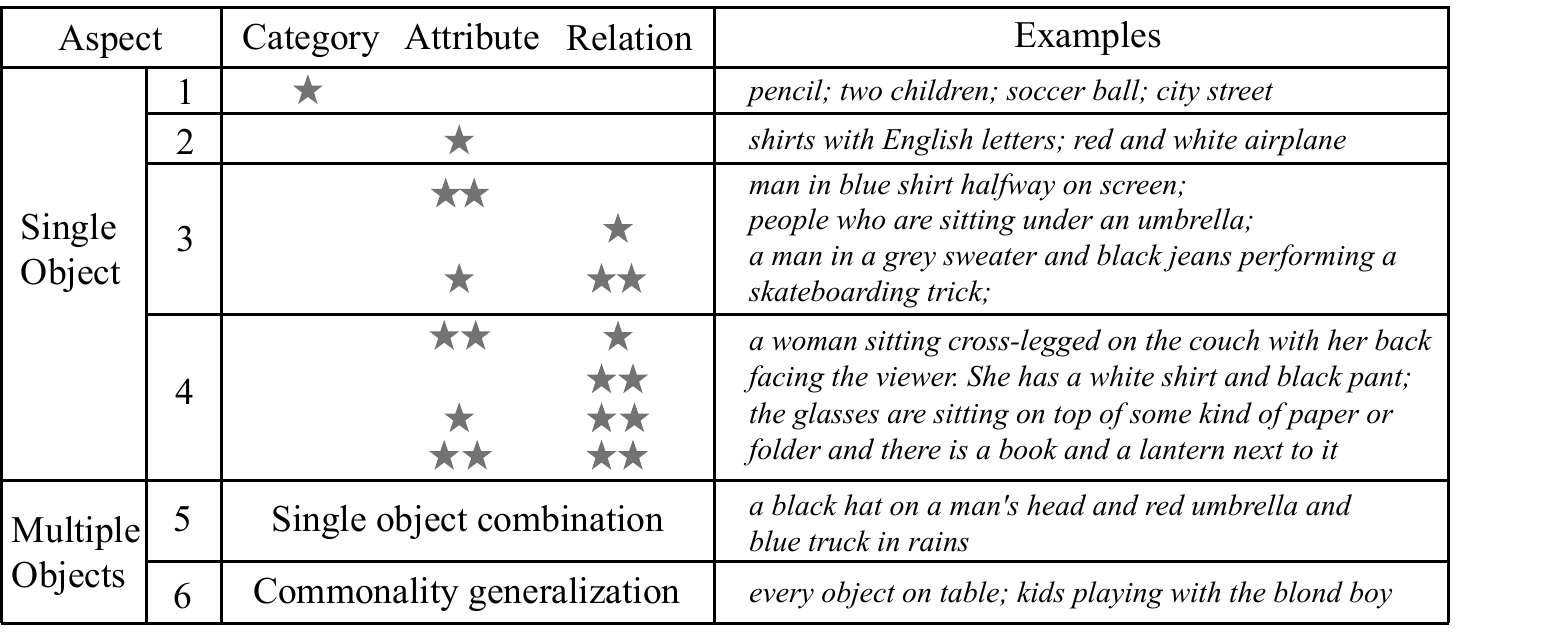}
    \label{tab:guideline}
    \vspace{-10pt}
\end{table}

{\bf Instruction Guideline}.
The instructions in InDET dataset describe objects from various perspectives. We observe that these descriptions all focus on object category, attribute, and relations, but with different emphasis extent. Based on expression complexity, we establish a guideline that divides all instructions into 6 groups. Each group reflects one level of emphasis on category, attribute, and relations. Table~\ref{tab:guideline} shows our guideline and examples. The first four groups are for single object and the last two groups are for multiple objects. In the first group (G1), there is only one single phrase to describe object category, which is similar to the traditional object detection task. From G2 to G4, more phrases are involved to describe the target object. For G5, we construct a spliced form to combine instructions from different single objects. In G6, the instruction is a general description of commonality between multiple objects. To this end, the instructions from G1 to G6 gradually introduces semantic understanding, visual language grounding, and logic reasoning for ROD. After guideline construction, we use LLaMA to assign each instruction into our groups by using in-context learning that let LLaMA to understand assigning principles and in-context assigning examples. The detailed usage of LLaMA for instruction assign is shown in Sec.~\ref{sec:supp-group}.

\begin{figure}
\centering
    \includegraphics[width=0.9\textwidth]{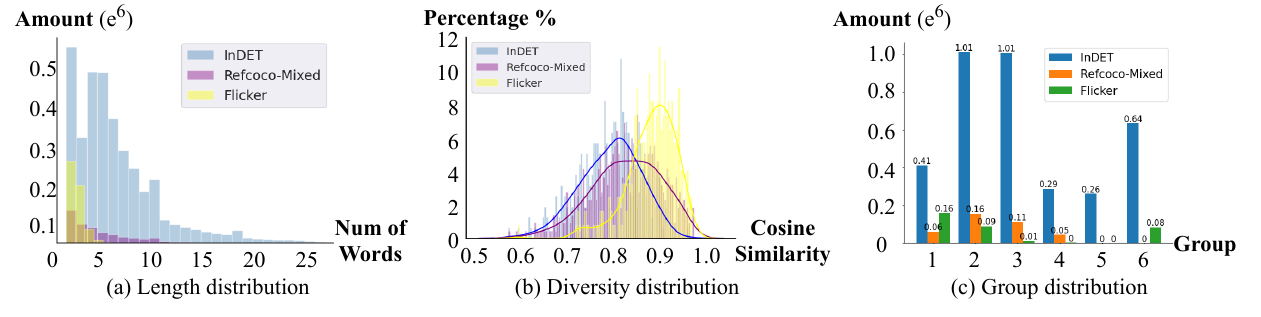}
    \vspace{-4mm}
    \caption{Dataset analysis of expression length, diversity and group distributions.}
    \label{fig:dataset_analysis}
    \vspace{-15pt}
\end{figure}

{\bf Instruction Length, Diversity, and Aspect Ratio Distributions}.
We analyze our InDET from the instruction length, diversity, and ratio distribution in our guideline groups. The RefCOCO and Flicker datasets are introduced for comparison. 
Fig.~\ref{fig:dataset_analysis}(a) shows the number of word distribution where the instruction of InDET contains more words than the other two datasets. Moreover, there are 100K instructions in our InDET consist of more than 10 words, while other datasets do not contain such informative expressions. In Fig.~\ref{fig:dataset_analysis}(b), we show diversity comparison where we use CLIP to map all instructions into a semantic space. Then, for the same target objects we compute average pairwise cosine similarity. The results show that our InDET contains lower value than other datasets, which indicates that our instructions are more diverse when describing the same target object. In Fig.~\ref{fig:dataset_analysis}(c), we show aspect ratio distribution of expressions assigned by our guideline. For existing datasets, user expressions commonly exist in G1 and G2. In contrast to the user expressions that seldom exist from G3 to G5 for Flicker, and seldom exist in G5 and G6 for RefCOCO, the instructions in our InDET exist normally in all groups. This distribution shows that our InDET is more effective to encompass common user intentions, especially for multiple objects. By leveraging our InDET, the ROD model becomes more practically applicable. 

\section{Referring Object Detection}
In this section, we illustrate our model design for ROD task. We notice that ROD shares little difference with visual grounding (VG). First, ROD produces uncertain number of object bbxs (i.e., 0, 1, or multiple) based on one input instruction, as shown in Fig.~\ref{fig:teaser}. Second, ROD supports abstract and summarized object descriptions (e.g., ``all objects on the table'') that do not clearly refer to specific objects such as ``bottle'', ``orange'', and ``knife''. As recent VG models~\citep{glipv2_nips22,gdino_arxiv23} require a one-by-one verification between visual objects and expression words, they are not able to execute such instructions. Motivated by the difference, we set up a conventional framework from DETR-based VG models~\citep{seqtr,UNINEXT}. Fig.~\ref{fig:baseline_model} shows an overview of our DROD model. We illustrate key steps as follows: 

\begin{wrapfigure}{R}{0.5\textwidth}
\vspace{-20pt}
\begin{center}
    \includegraphics[width=0.5\textwidth]{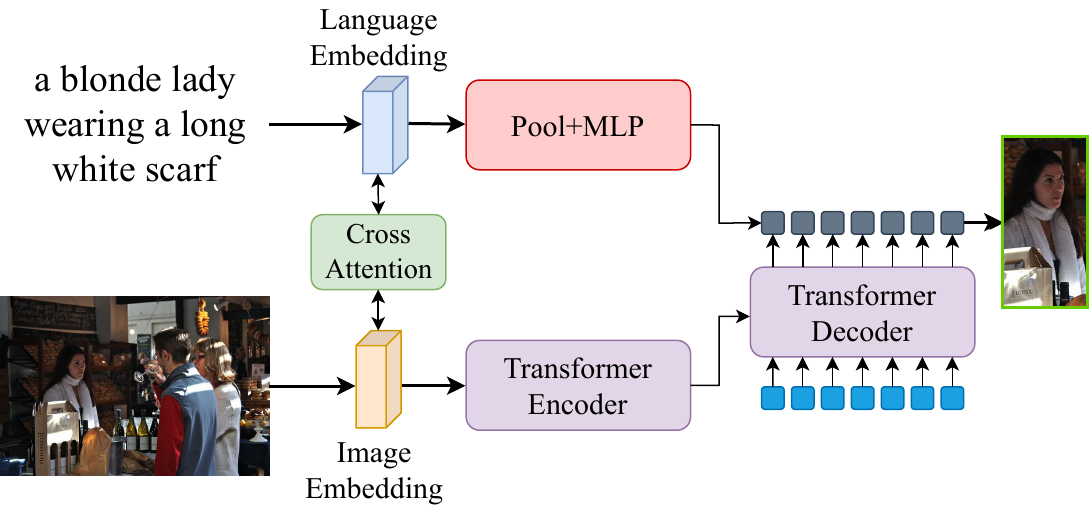}
    \vspace{-3mm}
    \caption{An overview of our diversified referring object detection (DROD) model.}
    \label{fig:baseline_model}
\end{center}
\vspace{-10pt}
\end{wrapfigure}

Given an image with text instruction, we use visual and text encoders~\citep{VIT,bert,ge2023soft} to obtain their embeddings. Then, we use a bi-directional cross attention module to perform multi-modality embedding fusion. For the fused visual embedding, we sent it to the transformer encoder and decoder structure~\citep{deformable} with $N$ learnable queries as position priors ~\citep{trackformer,ge2021revitalizing}. Then, the decoder produces $N$ instance proposals for further selection. For the fused text embedding, we pass it through a global average pooling and MLP for text2visual embedding space mapping. Finally, we use cosine similarity to match proposals and mapped text embedding. During the training stage, we use confidence loss and localization loss via supervised learning. During the inference stage, we select proposals whose matching scores are above a predefined threshold, which allows our model to produce arbitrary number of bbxs for diversified instruction execution. More details are shown in Sec.~\ref{sec:supp-model}.

\section{Experiments}
We evaluate the ROD performance on standard VG benchmarks (i.e., RefCOCO, RefCOCO+, and RefCOCOg) and our InDET dataset. As illustrated in Sec.~\ref{sec:analysis}, the images with marked objects of our InDET dataset are collected from existing datasets while the instructions are significantly enriched. We split the training and test set of InDET following RefCOCO/g/+ where the test set contains 6.5k images with an increased number of instructions to 315K. Moreover, these instructions are assigned to 6 groups according to our guideline. The performance on each group reflects how VG methods perform when processing different aspects of user instructions. The comparing methods in our experiments are from recent VG methods including MDETR~\citep{mdetr}, Grounding-DINO~\citep{gdino_arxiv23} and UNINEXT~\citep{UNINEXT}. Due to page limit, we show the evaluation results on our InDET, our InDET with shuffled expression, and standard benchmarks. Model training, and ablation studies on partial LLaVA finetuning and visual prompt selection are provided in Sec.~\ref{sec:supp-model} and Sec.~\ref{sec:supp-ablation}. We also provide visual comparison results of these methods in Sec.~\ref{sec:supp-visual}. A video demo showing the practical usage of our DROD model is in our webpage.





\begin{table}[h]
\scriptsize
\centering
\renewcommand{\tabcolsep}{1.2mm}
\renewcommand\arraystretch{1.1}
\setlength{\belowcaptionskip}{1pt}
\caption{Evaluation results on our InDET and shuffled InDET test sets. We show the object bbx average precision (AP) values ($\%$) of these two test sets with a slash (`/') separation.}
\label{tab:eval_on_InDET}
\vspace{-3mm}
\begin{tabular}{l|c|c|cccccc}
\hline
\multirow{2}{*}{Method} & \multirow{2}{*}{Backbone} & \multirow{2}{*}{AP} & \multicolumn{6}{c}{AP by Group}\\
 & & & G1 & G2 & G3 & G4 & G5 & G6 \\ \hline
 MDETR & ResNet101 & 34.86 / 31.21 & 47.44 / 46.61 & 46.79 / 42.55 & 34.14 / 28.13 & 23.22 / 16.86 & 25.91 / 23.52 & 28.17 / 23.66 \\
 G-DINO & SwinB & 35.96 / 30.43 & 47.10 / 45.91 & 47.17 / 42.56 & 35.29 / 27.28 & 26.84 / 18.46 & 27.95 / 23.74 & 27.61 / 23.57 \\
 UNINEXT & ResNet50 & 43.37 / 37.61 & 54.49 / 53.09 & 54.52 / 49.91 & 44.49 / 35.59 & 37.17 / 28.30 & 31.41 / 28.28 & 32.01 / 27.52 \\ \hline
 DROD (Ours) & ResNet50 & 62.24 / 53.78 & 67.14 / 65.08 & 67.34 / 61.56 & 60.89 / 48.82 & 55.10 / 41.50 & 70.15 / 64.64 & 74.22 / 67.11 \\
 DROD (Ours) & ViT-H & 66.90 / 57.32 & 72.53 / 69.79 & 72.47 / 65.44 & 66.42 / 52.50 & 59.86 / 46.01 & 73.34 / 67.82 & 75.92 / 68.73 \\ \hline
\end{tabular}
\end{table}

In our InDET test set, we compare our DROD model to other methods under the evaluation metric of object bbx average precision with a threshold of 0.5. On the other hand, we investigate whether these methods have truly comprehended the meaning of instruction, or they perform ROD only based on the key words (e.g., noun) without comprehending the whole expression. So we shuffle the InDET test set by randomly ordering the words in each instruction. We produce results of existing VG methods on our InDET test set without assuming object numbers in advance. For one method, if its performance drops more on the shuffled data, this method is shown to better comprehend the meaning of instruction.

Table~\ref{tab:eval_on_InDET} shows the evaluation results. Overall, UNINEXT achieves a higher AP than MDETR (i.e., 43.37 v.s. 34.86) in our InDET test set, while decreasing more than MDETR (i.e., 37.61 v.s. 31.21) in shuffled data. This indicates that UNINEXT is more effective than MDETR for ROD and better comprehends instruction meaning. Meanwhile, UNINEXT achieves a higher AP value than Grounding-DINO. In comparison, our DROD largely surpasses UNINEXT (62.24 v.s. 43.37) on the overall AP comparison, and using a VIT encoder further increases our performance. This indicates that our DROD is more effective to comprehend generalized instructions for ROD. Meanwhile, we observe that our performance drop is larger than UNINEXT (8.46 v.s. 5.76), which shows that our model better comprehends different expressions. Specifically for the results in each group, we notice that our performance drop is little in G1, and becomes larger from G2 to G4. This is because more and more words are introduced from G1 to G4 for object description. A random order gradually affects our model comprehension. For G5 and G6, we note that our method largely outperform other methods. The multi-object instructions incorporated in the dataset improves our performance.



\renewcommand{\tabcolsep}{1.8mm}
\begin{table}
\vspace{-1em}
\scriptsize
\centering
\renewcommand\arraystretch{1.1}
\setlength{\belowcaptionskip}{1pt}
\caption{Evaluation results on the RefCOCO/g/+ datasets. We follow evaluation protocols to report AP values ($\%$) of comparing methods. We use the notations "CC", "VG", "OI", "O365", "RIGame", for COCO, Visual Genome, OpenImage, Objects365, ReferItGame, respectively.}
\label{tab:eval_on_refcoco}
\vspace{-3mm}
\begin{threeparttable}
\begin{tabular}{l|c|c|ccc|ccc|cc}
\hline
\multirow{2}{*}{Method} & \multirow{2}{*}{Backbone} & \multirow{2}{*}{Data} & \multicolumn{3}{c|}{RefCOCO} & \multicolumn{3}{c|}{RefCOCO+} & \multicolumn{2}{c}{RefCOCOg}\\
 & & & val & testA & testB & val & testA & testB & val-u & test-u\\ \hline
 RefTR & ResNet101 & VG & 85.65 & 88.73 & 81.16 & 77.55 & 82.26 & 68.99 & 79.25 & 80.01 \\
 SeqTR & DarkNet53 & VG,RIGame,Flickr,RefC & 87.00 & 90.15 & 83.59 & 78.69 & 84.51 & 71.87 & 82.69 & 83.37 \\
 MDETR & ResNet101 & GoldG,CC,RefC & 86.75 & 89.58 & 81.41 & 79.52 & 84.09 & 70.62 & 81.64 & 80.89 \\
 G-DINO & SwinB & O365,CC,RefC,GoldG,etc & 83.95 & 87.79 & 79.16 & 72.91 & 80.91 & 62.96 & 76.98 & 76.76 \\
 UNINEXT & ResNet50 & O365,CC,RefC & 87.64 & 90.35 & 83.49 & 78.14 & 83.22 & 68.71 & 80.96 & 81.86 \\ 
 DROD (Ours) & ResNet50 & O365,CC,InDET & 88.92 & 90.86 & 85.57 & 78.27 & 83.39 & 71.04 & 83.01 & 82.91 \\ \hline
\end{tabular}
\end{threeparttable}
\vspace{-4mm}
\end{table}


Besides evaluating our InDET test set, we compare our DROD model with existing VG methods~\citep{seqtr,referTrans_nips21} on the standard VG benchmarks RefCOCO/g/+. Table~\ref{tab:eval_on_refcoco} shows the evaluation results. Overall, our DROD model achieves favorable performance on these datasets. This is because our DROD model utilizes InDET dataset where diversified instructions improve model generalizations. By using a conventional ROD model, we improve the VG performance from the data diversity perspective. 


\begin{figure}[t]
    \centering
    \includegraphics[width=0.9\textwidth]{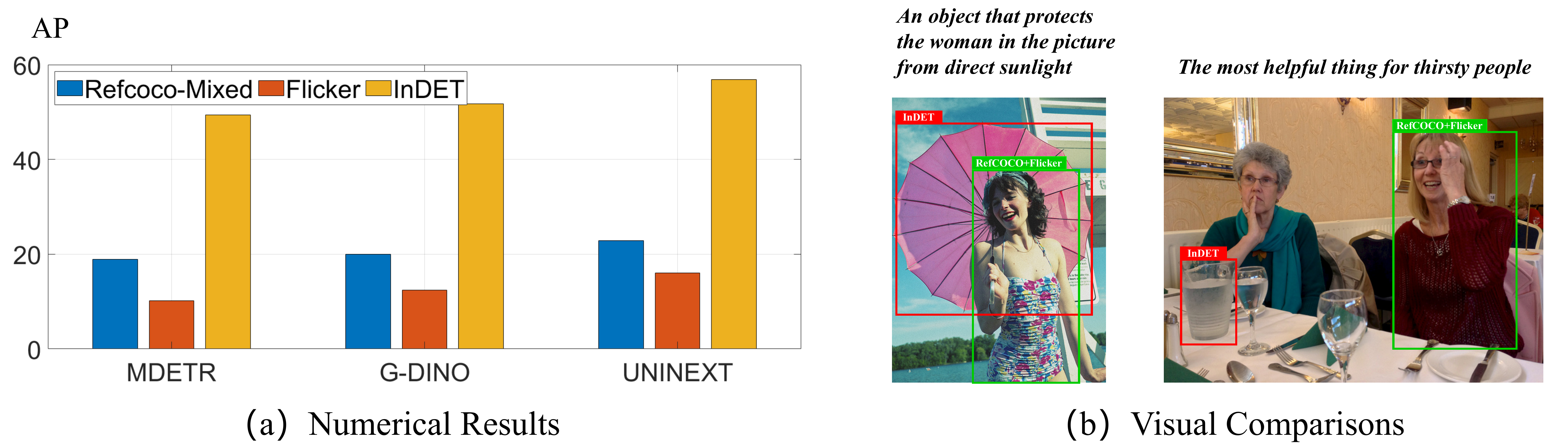}
    \vspace{-3mm}
    \caption{Our InDET dataset improves logic reasoning of ROD models. 
    In (a), existing models trained with our InDET dataset show superior results compared to other datasets. In (b), we show visual comparisons by using the same DROD model but with different training datasets.}
    \label{fig:logical}
    \vspace{-1em}
\end{figure}

In addition to the overall precision comparisons, we evaluate how our dataset improves logic reasoning and instruction comprehension of existing models. Specifically, we select 2k test samples from our InDET test dataset where logic reasoning on instructions is required for object detection. For each model (i.e., MDETR, G-DINO, or UNINEXT), we train it by using different datasets (i.e., RefCOCO, Flicker, or InDET) and show the performance comparison on our 2k test samples. Fig.~\ref{fig:logical}(a) shows the evaluation results where each model trained with our InDET dataset outperforms the same model trained with other datasets. In Fig.~\ref{fig:logical}(b), we show visual comparisons by using our DROD model but with different training sets. It shows that using original datasets, the model tends to ground keywords rather than preform multi-modal reasoning based on instructions. In comparison, by training with our InDET, the model well interprets instruction meaning and conduct logic reasoning across languages and visual images.


\section{Concluding Remarks}
We aim to push ROD into practical usage from a data-centric perspective. On one hand, we notice that current REC expressions are insufficient to encompass user detection intentions. On the other hand, foundation models have shown promising generalizations to simulate manual understanding and description abilities. To this end, we develop InstructDET that leverages foundation models to produce human-like expressions in REC, which tends to incorporate common user intentions into ROD training. As a result, our DROD model achieves favorable performance compared to existing VG methods. In the future, we can combine our method with open-set object detectors to fully explore in-the-wild images (e.g., Internet images) for comprehensive user expression generation. We expect our DROD model to generalize as much as existing foundation models, and thus take a huge step towards completely solving ROD task.  

\section*{Acknowledgement}
This paper is supported by the National Natural Science Foundation of China under Grants (62073245, 62173248,62233013). Shanghai Science and Technology Innovation Action Plan (22511104900). Shanghai Municipal Science and Technology Major Project (2021SHZDZX0100) and the Fundamental Research Funds for the Central Universities.

{\small
\normalem
\bibliography{iclr2024_conference}

\begin{thebibliography}{54}
\providecommand{\natexlab}[1]{#1}
\providecommand{\url}[1]{\texttt{#1}}
\expandafter\ifx\csname urlstyle\endcsname\relax
  \providecommand{\doi}[1]{doi: #1}\else
  \providecommand{\doi}{doi: \begingroup \urlstyle{rm}\Url}\fi

\bibitem[Akbari et~al.(2019)Akbari, Karaman, Bhargava, Chen, Vondrick, and Chang]{multi-level_IPG}
Hassan Akbari, Svebor Karaman, Surabhi Bhargava, Brian Chen, Carl Vondrick, and Shih-Fu Chang.
\newblock Multi-level multimodal common semantic space for image-phrase grounding.
\newblock In \emph{IEEE/CVF Conference on Computer Vision and Pattern Recognition}, 2019.

\bibitem[Carion et~al.(2020)Carion, Massa, Synnaeve, Usunier, Kirillov, and Zagoruyko]{DETR}
Nicolas Carion, Francisco Massa, Gabriel Synnaeve, Nicolas Usunier, Alexander Kirillov, and Sergey Zagoruyko.
\newblock End-to-end object detection with transformers.
\newblock In \emph{European Conference on Computer Vision}, 2020.

\bibitem[Chen et~al.(2020)Chen, Wang, Ma, Wong, and Wu]{cops-ref}
Zhenfang Chen, Peng Wang, Lin Ma, Kwan-Yee~K Wong, and Qi~Wu.
\newblock Cops-ref: A new dataset and task on compositional referring expression comprehension.
\newblock In \emph{IEEE/CVF Conference on Computer Vision and Pattern Recognition}, 2020.

\bibitem[Chen et~al.(2021)Chen, Mao, Wu, Wong, Tenenbaum, and Gan]{groundobj_2021_iclr}
Zhenfang Chen, Jiayuan Mao, Jiajun Wu, Kwan-Yee~Kenneth Wong, Joshua~B. Tenenbaum, and Chuang Gan.
\newblock Grounding physical concepts of objects and events through dynamic visual reasoning.
\newblock In \emph{International Conference on Learning Representations}, 2021.

\bibitem[Deng et~al.(2021)Deng, Yang, Chen, Zhou, and Li]{transVG_iccv21}
Jiajun Deng, Zhengyou Yang, Tianlang Chen, Wengang Zhou, and Houqiang Li.
\newblock Transvg: End-to-end visual grounding with transformers.
\newblock In \emph{IEEE/CVF International Conference on Computer Vision}, 2021.

\bibitem[Devlin et~al.(2018)Devlin, Chang, Lee, and Toutanova]{bert}
Jacob Devlin, Ming-Wei Chang, Kenton Lee, and Kristina Toutanova.
\newblock Bert: Pre-training of deep bidirectional transformers for language understanding.
\newblock \emph{ArXiv preprint arXiv:1810.04805}, 2018.

\bibitem[Dong et~al.(2023)Dong, Li, Dai, Zheng, Wu, Chang, Sun, Xu, Li, and Sui]{icl}
Qingxiu Dong, Lei Li, Damai Dai, Ce~Zheng, Zhiyong Wu, Baobao Chang, Xu~Sun, Jingjing Xu, Lei Li, and Zhifang Sui.
\newblock A survey on in-context learning.
\newblock \emph{ArXiv preprint arXiv:2301.00234}, 2023.

\bibitem[Dosovitskiy et~al.(2021)Dosovitskiy, Beyer, Kolesnikov, Weissenborn, Zhai, Unterthiner, Dehghani, Minderer, Heigold, Gelly, et~al.]{VIT}
Alexey Dosovitskiy, Lucas Beyer, Alexander Kolesnikov, Dirk Weissenborn, Xiaohua Zhai, Thomas Unterthiner, Mostafa Dehghani, Matthias Minderer, Georg Heigold, Sylvain Gelly, et~al.
\newblock An image is worth 16x16 words: Transformers for image recognition at scale.
\newblock In \emph{International Conference on Learning Representations}, 2021.

\bibitem[Ester et~al.(1996)Ester, Kriegel, Sander, Xu, et~al.]{DBSCAN}
Martin Ester, Hans-Peter Kriegel, J{\"o}rg Sander, Xiaowei Xu, et~al.
\newblock A density-based algorithm for discovering clusters in large spatial databases with noise.
\newblock In \emph{Special Interest Group on Knowledge Discovery and Data Mining}, 1996.

\bibitem[Gao et~al.(2023)Gao, Uzkent, Shen, Huang, and Jin]{gvlm_2023_iclr}
Shangqian Gao, Burak Uzkent, Yilin Shen, Heng Huang, and Hongxia Jin.
\newblock Learning to jointly share and prune weights for grounding based vision and language models.
\newblock In \emph{International Conference on Learning Representations}, 2023.

\bibitem[Ge et~al.(2021)Ge, Liang, Song, Jiao, Wang, and Luo]{ge2021revitalizing}
Chongjian Ge, Youwei Liang, Yibing Song, Jianbo Jiao, Jue Wang, and Ping Luo.
\newblock Revitalizing cnn attentions via transformers in self-supervised visual representation learning.
\newblock In \emph{Advances in Neural Information Processing Systems}, 2021.

\bibitem[Ge et~al.(2023)Ge, Wang, Tong, Chen, Song, and Luo]{ge2023soft}
Chongjian Ge, Jiangliu Wang, Zhan Tong, Shoufa Chen, Yibing Song, and Ping Luo.
\newblock Soft neighbors are positive supporters in contrastive visual representation learning.
\newblock In \emph{International Conference on Learning Representations}, 2023.

\bibitem[Gupta \& Kembhavi(2023)Gupta and Kembhavi]{VISPROG_2023_CVPR}
Tanmay Gupta and Aniruddha Kembhavi.
\newblock Visual programming: Compositional visual reasoning without training.
\newblock In \emph{IEEE/CVF Conference on Computer Vision and Pattern Recognition}, 2023.

\bibitem[He et~al.(2016)He, Zhang, Ren, and Sun]{resnet}
Kaiming He, Xiangyu Zhang, Shaoqing Ren, and Jian Sun.
\newblock Deep residual learning for image recognition.
\newblock In \emph{IEEE/CVF Conference on Computer Vision and Pattern Recognition}, 2016.

\bibitem[Hudson \& Manning(2018)Hudson and Manning]{comatt_iclr18}
Drew Hudson and Christopher Manning.
\newblock Compositional attention networks for machine reasoning.
\newblock In \emph{International Conference on Learning Representations}, 2018.

\bibitem[Kamath et~al.(2021)Kamath, Singh, LeCun, Misra, Synnaeve, and Carion]{mdetr}
Aishwarya Kamath, Mannat Singh, Yann LeCun, Ishan Misra, Gabriel Synnaeve, and Nicolas Carion.
\newblock Mdetr--modulated detection for end-to-end multi-modal understanding.
\newblock In \emph{IEEE/CVF International Conference on Computer Vision}, 2021.

\bibitem[Kazemzadeh et~al.(2014)Kazemzadeh, Ordonez, Matten, and Berg]{referitgame}
Sahar Kazemzadeh, Vicente Ordonez, Mark Matten, and Tamara Berg.
\newblock Referitgame: Referring to objects in photographs of natural scenes.
\newblock In \emph{Conference on Empirical Methods in Natural Language Processing}, 2014.

\bibitem[Kebe et~al.(2021)Kebe, Higgins, Jenkins, Darvish, Sachdeva, Barron, Winder, Engel, Raff, Ferraro, and Matuszek]{goldg-nips21}
Gaoussou~Youssouf Kebe, Padraig Higgins, Patrick Jenkins, Kasra Darvish, Rishabh Sachdeva, Ryan Barron, John Winder, Donald Engel, Edward Raff, Francis Ferraro, and Cynthia Matuszek.
\newblock A spoken language dataset of descriptions for speech-based grounded language learning.
\newblock In \emph{Advances on Neural Information Processing Systems Datasets and Benchmarks Track}, 2021.

\bibitem[Kirillov et~al.(2023)Kirillov, Mintun, Ravi, Mao, Rolland, Gustafson, Xiao, Whitehead, Berg, Lo, et~al.]{SAM}
Alexander Kirillov, Eric Mintun, Nikhila Ravi, Hanzi Mao, Chloe Rolland, Laura Gustafson, Tete Xiao, Spencer Whitehead, Alexander~C Berg, Wan-Yen Lo, et~al.
\newblock Segment anything.
\newblock \emph{ArXiv preprint arXiv:2304.02643}, 2023.

\bibitem[Kojima et~al.(2023)Kojima, Averbuch-Elor, and Artzi]{PGTP}
Noriyuki Kojima, Hadar Averbuch-Elor, and Yoav Artzi.
\newblock A joint study of phrase grounding and task performance in vision and language models.
\newblock \emph{ArXiv preprint arXiv:2309.02691}, 2023.

\bibitem[Krishna et~al.(2017)Krishna, Zhu, Groth, Johnson, Hata, Kravitz, Chen, Kalantidis, Li, Shamma, et~al.]{krishna2017visual}
Ranjay Krishna, Yuke Zhu, Oliver Groth, Justin Johnson, Kenji Hata, Joshua Kravitz, Stephanie Chen, Yannis Kalantidis, Li-Jia Li, David~A Shamma, et~al.
\newblock Visual genome: Connecting language and vision using crowdsourced dense image annotations.
\newblock \emph{International Journal of Computer Vision}, 2017.

\bibitem[Kuznetsova et~al.(2020)Kuznetsova, Rom, Alldrin, Uijlings, Krasin, Pont-Tuset, Kamali, Popov, Malloci, Kolesnikov, Duerig, and Ferrari]{openimg-ijcv20}
Alina Kuznetsova, Hassan Rom, Neil Alldrin, Jasper Uijlings, Ivan Krasin, Jordi Pont-Tuset, Shahab Kamali, Stefan Popov, Matteo Malloci, Alexander Kolesnikov, Tom Duerig, and Vittorio Ferrari.
\newblock The open images dataset v4: Unified image classification, object detection, and visual relationship detection at scale.
\newblock \emph{International Journal of Computer Vision}, 2020.

\bibitem[Li et~al.(2023{\natexlab{a}})Li, Zhang, Chen, Wang, Yang, and Liu]{otter}
Bo~Li, Yuanhan Zhang, Liangyu Chen, Jinghao Wang, Jingkang Yang, and Ziwei Liu.
\newblock Otter: A multi-modal model with in-context instruction tuning.
\newblock \emph{ArXiv preprint arXiv:2305.03726}, 2023{\natexlab{a}}.

\bibitem[Li et~al.(2023{\natexlab{b}})Li, Li, Savarese, and Hoi]{blip-2}
Junnan Li, Dongxu Li, Silvio Savarese, and Steven Hoi.
\newblock Blip-2: Bootstrapping language-image pre-training with frozen image encoders and large language models.
\newblock \emph{ArXiv preprint arXiv:2301.12597}, 2023{\natexlab{b}}.

\bibitem[Li et~al.(2022{\natexlab{a}})Li, Zhang, Zhang, Yang, Li, Zhong, Wang, Yuan, Zhang, Hwang, Chang, and Gao]{Li_2022_CVPR}
Liunian~Harold Li, Pengchuan Zhang, Haotian Zhang, Jianwei Yang, Chunyuan Li, Yiwu Zhong, Lijuan Wang, Lu~Yuan, Lei Zhang, Jenq-Neng Hwang, Kai-Wei Chang, and Jianfeng Gao.
\newblock Grounded language-image pre-training.
\newblock In \emph{IEEE/CVF Conference on Computer Vision and Pattern Recognition}, 2022{\natexlab{a}}.

\bibitem[Li \& Sigal(2021)Li and Sigal]{referTrans_nips21}
Muchen Li and Leonid Sigal.
\newblock Referring transformer: A one-step approach to multi-task visual grounding.
\newblock In \emph{Advances in Neural Information Processing Systems}, 2021.

\bibitem[Li et~al.(2022{\natexlab{b}})Li, Thickstun, Gulrajani, Liang, and Hashimoto]{diffusion-lm}
Xiang Li, John Thickstun, Ishaan Gulrajani, Percy~S Liang, and Tatsunori~B Hashimoto.
\newblock Diffusion-lm improves controllable text generation.
\newblock \emph{Advances in Neural Information Processing Systems}, 2022{\natexlab{b}}.

\bibitem[Lin et~al.(2014)Lin, Maire, Belongie, Hays, Perona, Ramanan, Doll{\'a}r, and Zitnick]{mscoco}
Tsung-Yi Lin, Michael Maire, Serge Belongie, James Hays, Pietro Perona, Deva Ramanan, Piotr Doll{\'a}r, and C~Lawrence Zitnick.
\newblock Microsoft coco: Common objects in context.
\newblock In \emph{European Conference on Computer Vision}, 2014.

\bibitem[Liu et~al.(2023{\natexlab{a}})Liu, Li, Wu, and Lee]{LLAVA}
Haotian Liu, Chunyuan Li, Qingyang Wu, and Yong~Jae Lee.
\newblock Visual instruction tuning.
\newblock \emph{ArXiv preprint arXiv:2304.08485}, 2023{\natexlab{a}}.

\bibitem[Liu et~al.(2023{\natexlab{b}})Liu, Ding, Cai, Zhang, Satzoda, Mahadevan, and Manmatha]{polyformer}
Jiang Liu, Hui Ding, Zhaowei Cai, Yuting Zhang, Ravi~Kumar Satzoda, Vijay Mahadevan, and R~Manmatha.
\newblock Polyformer: Referring image segmentation as sequential polygon generation.
\newblock In \emph{Proceedings of the IEEE/CVF Conference on Computer Vision and Pattern Recognition}, pp.\  18653--18663, 2023{\natexlab{b}}.

\bibitem[Liu et~al.(2023{\natexlab{c}})Liu, Zeng, Ren, Li, Zhang, Yang, Li, Yang, Su, Zhu, and Zhang]{gdino_arxiv23}
Shilong Liu, Zhaoyang Zeng, Tianhe Ren, Feng Li, Hao Zhang, Jie Yang, Chunyuan Li, Jianwei Yang, Hang Su, Jun Zhu, and Lei Zhang.
\newblock Grounding dino: Marrying dino with grounded pre-training for open-set object detection.
\newblock In \emph{ArXiv preprint arXiv:2303.05499}, 2023{\natexlab{c}}.

\bibitem[Loshchilov \& Hutter(2019)Loshchilov and Hutter]{adamw}
Ilya Loshchilov and Frank Hutter.
\newblock Decoupled weight decay regularization.
\newblock In \emph{International Conference on Learning Representations}, 2019.

\bibitem[Mao et~al.(2016)Mao, Huang, Toshev, Camburu, Yuille, and Murphy]{refcocog}
Junhua Mao, Jonathan Huang, Alexander Toshev, Oana Camburu, Alan~L Yuille, and Kevin Murphy.
\newblock Generation and comprehension of unambiguous object descriptions.
\newblock In \emph{IEEE/CVF Conference on Computer Vision and Pattern Recognition}, 2016.

\bibitem[Meinhardt et~al.(2022)Meinhardt, Kirillov, Leal-Taixe, and Feichtenhofer]{trackformer}
Tim Meinhardt, Alexander Kirillov, Laura Leal-Taixe, and Christoph Feichtenhofer.
\newblock Trackformer: Multi-object tracking with transformers.
\newblock In \emph{IEEE/CVF Conference on Computer Vision and Pattern Recognition}, 2022.

\bibitem[OpenAI(2023)]{gpt-4}
OpenAI.
\newblock Gpt-4 technical report.
\newblock 2023.

\bibitem[Ouyang et~al.(2022)Ouyang, Wu, Jiang, Almeida, Wainwright, Mishkin, Zhang, Agarwal, Slama, Ray, et~al.]{instructgpt}
Long Ouyang, Jeffrey Wu, Xu~Jiang, Diogo Almeida, Carroll Wainwright, Pamela Mishkin, Chong Zhang, Sandhini Agarwal, Katarina Slama, Alex Ray, et~al.
\newblock Training language models to follow instructions with human feedback.
\newblock In \emph{Advances in Neural Information Processing Systems}, 2022.

\bibitem[Plummer et~al.(2022)Plummer, Dryden, Frost, Hoefler, and Saenko]{npas_iclr22}
Bryan Plummer, Nikoli Dryden, Julius Frost, Torsten Hoefler, and Kate Saenko.
\newblock Neural parameter allocation search.
\newblock In \emph{International Conference on Learning Representations}, 2022.

\bibitem[Plummer et~al.(2015)Plummer, Wang, Cervantes, Caicedo, Hockenmaier, and Lazebnik]{flickr30k}
Bryan~A Plummer, Liwei Wang, Chris~M Cervantes, Juan~C Caicedo, Julia Hockenmaier, and Svetlana Lazebnik.
\newblock Flickr30k entities: Collecting region-to-phrase correspondences for richer image-to-sentence models.
\newblock In \emph{IEEE/CVF International Conference on Computer Vision}, 2015.

\bibitem[Radford et~al.(2021)Radford, Kim, Hallacy, Ramesh, Goh, Agarwal, Sastry, Askell, Mishkin, Clark, Krueger, and Sutskever]{clip}
Alec Radford, Jong~Wook Kim, Chris Hallacy, Aditya Ramesh, Gabriel Goh, Sandhini Agarwal, Girish Sastry, Amanda Askell, Pamela Mishkin, Jack Clark, Gretchen Krueger, and Ilya Sutskever.
\newblock Learning transferable visual models from natural language supervision.
\newblock In \emph{International Conference on Machine Learning}, 2021.

\bibitem[Shao et~al.(2019)Shao, Li, Zhang, Peng, Yu, Zhang, Li, and Sun]{objects365}
Shuai Shao, Zeming Li, Tianyuan Zhang, Chao Peng, Gang Yu, Xiangyu Zhang, Jing Li, and Jian Sun.
\newblock Objects365: A large-scale, high-quality dataset for object detection.
\newblock In \emph{IEEE/CVF International Conference on Computer Vision}, 2019.

\bibitem[Shtedritski et~al.(2023)Shtedritski, Rupprecht, and Vedaldi]{red_circle}
Aleksandar Shtedritski, Christian Rupprecht, and Andrea Vedaldi.
\newblock What does clip know about a red circle? visual prompt engineering for vlms.
\newblock \emph{ArXiv preprint arXiv:2304.06712}, 2023.

\bibitem[Su et~al.(2023)Su, Miao, Dou, Wang, Qiao, Li, and Li]{Su_2023_CVPR}
Wei Su, Peihan Miao, Huanzhang Dou, Gaoang Wang, Liang Qiao, Zheyang Li, and Xi~Li.
\newblock Language adaptive weight generation for multi-task visual grounding.
\newblock In \emph{IEEE/CVF Conference on Computer Vision and Pattern Recognition}, 2023.

\bibitem[Su et~al.(2020)Su, Zhu, Cao, Li, Lu, Wei, and Dai]{vlbert_iclr20}
Weijie Su, Xizhou Zhu, Yue Cao, Bin Li, Lewei Lu, Furu Wei, and Jifeng Dai.
\newblock Vl-bert: Pre-training of generic visual-linguistic representations.
\newblock In \emph{International Conference on Learning Representations}, 2020.

\bibitem[Touvron et~al.(2023)Touvron, Lavril, Izacard, Martinet, Lachaux, Lacroix, Rozière, Goyal, Hambro, Azhar, Rodriguez, Joulin, Grave, and Lample]{LLaMA}
Hugo Touvron, Thibaut Lavril, Gautier Izacard, Xavier Martinet, Marie-Anne Lachaux, Timothée Lacroix, Baptiste Rozière, Naman Goyal, Eric Hambro, Faisal Azhar, Aurelien Rodriguez, Armand Joulin, Edouard Grave, and Guillaume Lample.
\newblock Llama: Open and efficient foundation language models.
\newblock \emph{ArXiv preprint arXiv:2302.13971}, 2023.

\bibitem[Wang et~al.(2022)Wang, Roberts, Hesslow, Le~Scao, Chung, Beltagy, Launay, and Raffel]{llm-generalization}
Thomas Wang, Adam Roberts, Daniel Hesslow, Teven Le~Scao, Hyung~Won Chung, Iz~Beltagy, Julien Launay, and Colin Raffel.
\newblock What language model architecture and pretraining objective works best for zero-shot generalization?
\newblock In \emph{International Conference on Machine Learning}, 2022.

\bibitem[Wang et~al.(2021)Wang, Xu, Wang, Shen, Cheng, Shen, and Xia]{VIST}
Yuqing Wang, Zhaoliang Xu, Xinlong Wang, Chunhua Shen, Baoshan Cheng, Hao Shen, and Huaxia Xia.
\newblock End-to-end video instance segmentation with transformers.
\newblock In \emph{IEEE/CVF Conference on Computer Vision and Pattern Recognition}, 2021.

\bibitem[Wu et~al.(2023)Wu, Yin, Qi, Wang, Tang, and Duan]{visualchatgpt}
Chenfei Wu, Shengming Yin, Weizhen Qi, Xiaodong Wang, Zecheng Tang, and Nan Duan.
\newblock Visual chatgpt: Talking, drawing and editing with visual foundation models.
\newblock \emph{ArXiv preprint arXiv:2303.04671}, 2023.

\bibitem[Yan et~al.(2023)Yan, Jiang, Wu, Wang, Luo, Yuan, and Lu]{UNINEXT}
Bin Yan, Yi~Jiang, Jiannan Wu, Dong Wang, Ping Luo, Zehuan Yuan, and Huchuan Lu.
\newblock Universal instance perception as object discovery and retrieval.
\newblock In \emph{IEEE/CVF Conference on Computer Vision and Pattern Recognition}, 2023.

\bibitem[Yang et~al.(2023)Yang, Wang, Li, Wang, and Yang]{FGVP}
Lingfeng Yang, Yueze Wang, Xiang Li, Xinlong Wang, and Jian Yang.
\newblock Fine-grained visual prompting.
\newblock \emph{ArXiv preprint arXiv:2306.04356}, 2023.

\bibitem[Yu et~al.(2016)Yu, Poirson, Yang, Berg, and Berg]{refcoco}
Licheng Yu, Patrick Poirson, Shan Yang, Alexander~C Berg, and Tamara~L Berg.
\newblock Modeling context in referring expressions.
\newblock In \emph{European Conference on Computer Vision}, 2016.

\bibitem[Zhang et~al.(2022)Zhang, Zhang, Hu, Chen, Li, Dai, Wang, Yuan, Hwang, and Gao]{glipv2_nips22}
Haotian Zhang, Pengchuan Zhang, Xiaowei Hu, Yen-Chun Chen, Liunian Li, Xiyang Dai, Lijuan Wang, Lu~Yuan, Jenq-Neng Hwang, and Jianfeng Gao.
\newblock Glipv2: Unifying localization and vision-language understanding.
\newblock In \emph{Advances in Neural Information Processing Systems}, 2022.

\bibitem[Zhou et~al.(2022)Zhou, Sch{\"a}rli, Hou, Wei, Scales, Wang, Schuurmans, Cui, Bousquet, Le, et~al.]{llm-reasoning}
Denny Zhou, Nathanael Sch{\"a}rli, Le~Hou, Jason Wei, Nathan Scales, Xuezhi Wang, Dale Schuurmans, Claire Cui, Olivier Bousquet, Quoc Le, et~al.
\newblock Least-to-most prompting enables complex reasoning in large language models.
\newblock \emph{ArXiv preprint arXiv:2205.10625}, 2022.

\bibitem[Zhu et~al.(2022)Zhu, Zhou, Shen, Luo, Pan, Lin, Chen, Cao, Sun, and Ji]{seqtr}
Chaoyang Zhu, Yiyi Zhou, Yunhang Shen, Gen Luo, Xingjia Pan, Mingbao Lin, Chao Chen, Liujuan Cao, Xiaoshuai Sun, and Rongrong Ji.
\newblock Seqtr: A simple yet universal network for visual grounding.
\newblock In \emph{European Conference on Computer Vision}, 2022.

\bibitem[Zhu et~al.(2020)Zhu, Su, Lu, Li, Wang, and Dai]{deformable}
Xizhou Zhu, Weijie Su, Lewei Lu, Bin Li, Xiaogang Wang, and Jifeng Dai.
\newblock Deformable detr: Deformable transformers for end-to-end object detection.
\newblock \emph{ArXiv preprint arXiv:2010.04159}, 2020.

\end{thebibliography}
\bibliographystyle{iclr2024_conference}}



\clearpage
\appendix
\section*{Appendix Overview}
We provide an overview to present a clear understanding of this section. 

\begin{itemize}
 \item In Sec.~\ref{sec:supp-visual}, we present visual comparisons of recent VG methods under various instructions. 
 \item In Sec.~\ref{sec:supp-multi-modality} and Sec.~\ref{sec:supp-single-modality}, we describe the process of generating expressions from foundation models in both global prompt and local prompt pipelines, respectively. 
\item In Sec.~\ref{sec:supp-vprompt}, we detail the aspects of visual prompting and visual-textual matching. 
\item In Sec.~\ref{sec:supp-multiobj}, we explain our approach to designing prompts that enable LLaVA to extract commonalities among multiple objects for summarized expression generation. 
\item In Sec.~\ref{sec:supp-group}, we discuss our strategy for designing prompts that allow LLaMA to assign generated instructions to our predefined groups. 
\item In Sec.~\ref{sec:supp-model}, we provide an overview of our model architecture and implementation specifics. 
\item In Sec.~\ref{sec:supp-ablation}, we present ablation studies investigating LLaVA fine-tuning in local prompt pipeline and visual prompting selection. 
\item In Sec.~\ref{sec:supp-syn-rewrite}, we provide the prompt we used in post processing.
\item In Sec.~\ref{sec:supp-comparison}, we add supplementary evaluation results to Table.~\ref{tab:eval_on_InDET} and Table.~\ref{tab:eval_on_refcoco}.
\item In Sec.~\ref{sec:supp-time-consumption}, we add the time consumption of main steps in our instruction generation procedure.
\end{itemize}


\section{Visual Comparison Results}\label{sec:supp-visual}

\begin{figure}[h]
    \centering
    \includegraphics[width=0.8\textwidth]{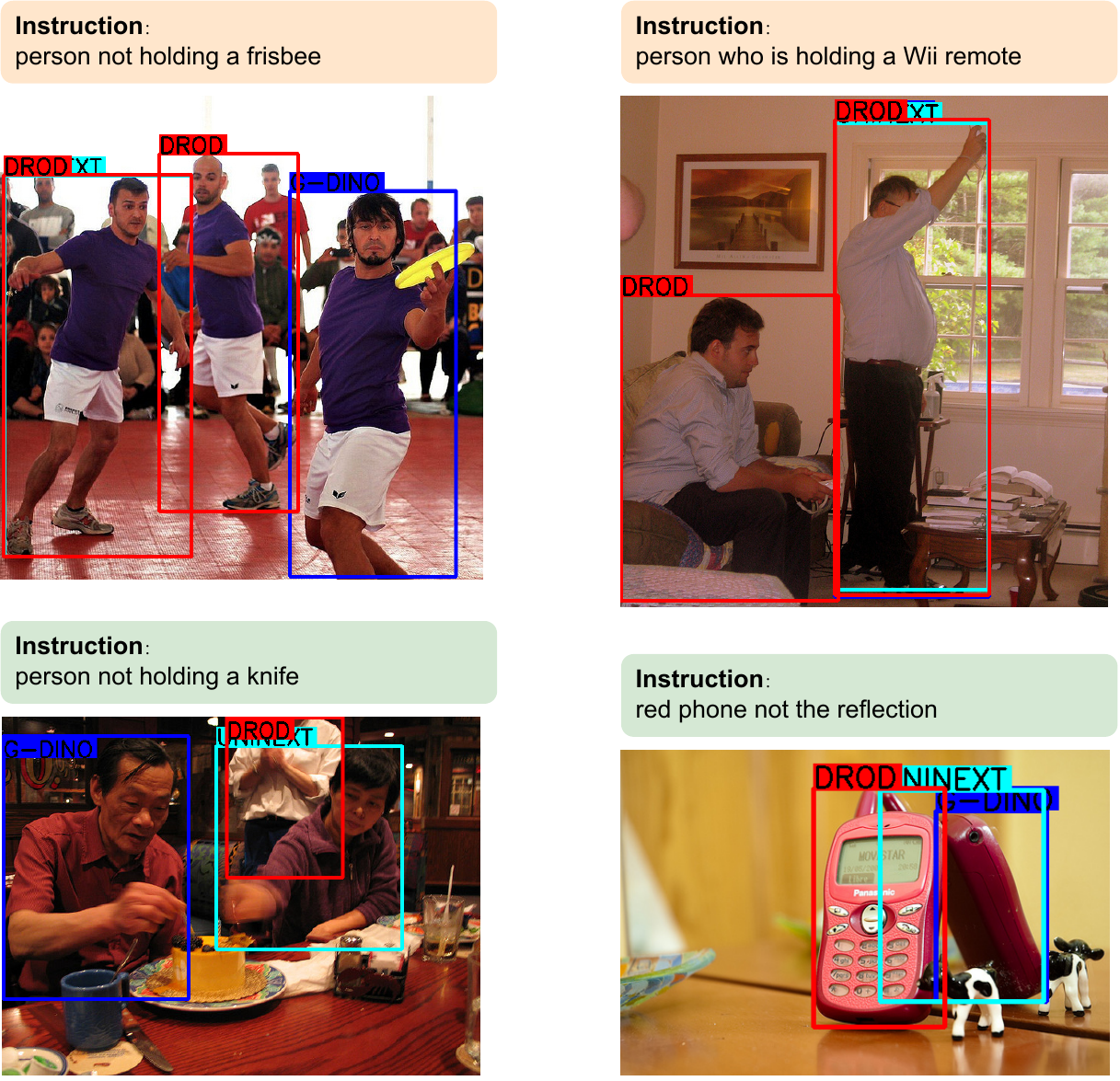}
    \caption{Visual comparison results. Our DROD well executes detection instruction compared to Grounding-DINO and UNINEXT.}
    \label{fig:visual_comparison_results_part2}
\end{figure}

\begin{figure}[h]
    \centering
    \includegraphics[width=0.8\textwidth]{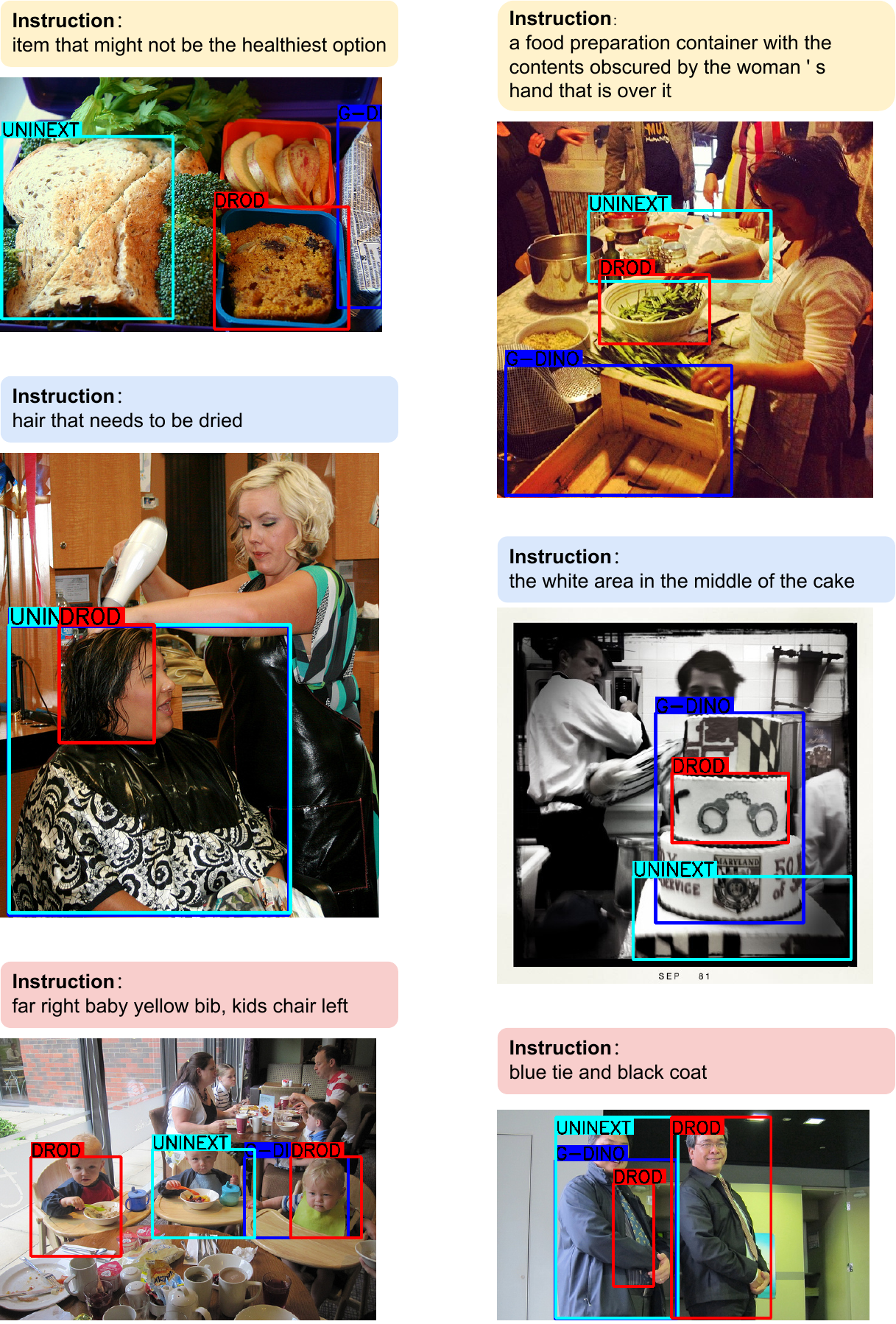}
    \caption{Visual comparison results. Our DROD well executes detection instruction compared to Grounding-DINO and UNINEXT.}
    \label{fig:visual_comparison_results_part1}
\end{figure}

\clearpage

\section{Expression Generation via LOCAL PROMPT PIPELINE}\label{sec:supp-multi-modality}

\subsection{Prompt and Generation Display}
\begin{figure}[h]
    \centering
    \includegraphics[width=\textwidth]{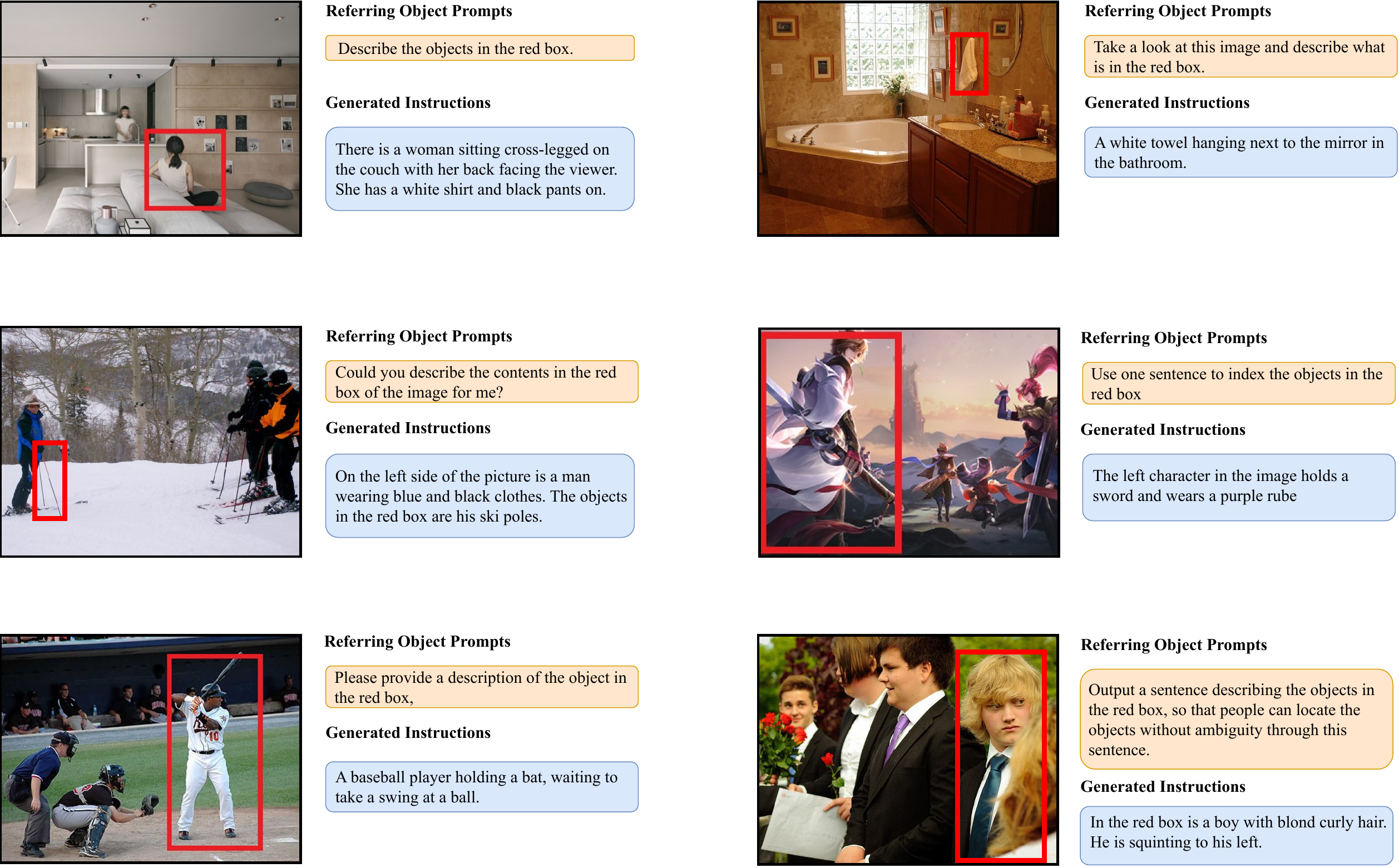}
    \caption{Examples of text prompts, images with marked object bboxs, and LLaVA outputs.}
    \label{fig:LMM_res}
\end{figure}

We show the text prompts we use below in Table~\ref{tab:table_lmm_instruction}. For the LLaVA input, We randomly select one text prompt together with one image marked with object bbox. The inputs and samples generated from LLaVA are shown in Fig.~\ref{fig:LMM_res}.

\begin{table}[h]
\begin{tcolorbox}[colback=gray!10,
                  colframe=black,
                  width=14cm,
                  arc=3mm, auto outer arc,
                 ]

\begin{itemize}
\item  ``Describe the objects in the red box."

\item   ``Take a look at this image and describe What's in the red box."
\item   ``Please provide a description of the object in the red box."

\item   ``Could you describe the contents in the red box of the image for me?"
\item   ``Use one sentence to index the objects in the red box."

\item   ``Output a sentence describing the objects in the red box, so that people can locate the objects without ambiguity through this sentence."

\item   ``Look carefully at the objects in the red box and describe them in one sentence to distinguish them from other objects."

\end{itemize}
\end{tcolorbox}
\caption{Text prompts we use for LLaVA inputs.}
\label{tab:table_lmm_instruction}
\end{table}

\subsection{Fine-tuning Settings}

\paragraph{Dataset Construction}
We employed the entire dataset from RefCOCO/+/g for our fine-tuning endeavors, which primarily comprises triplets of images, bounding boxes, and instructions. The detection boxes are overlaid onto the images, serving as the visual input for LLAVA, while the instructions from the dataset are used as supervised outputs. The prompts are randomly selected from Table~\ref{tab:table_lmm_instruction}. The overall dataset consists of 386k instructions paired with 66k images. We chose not to utilize additional REC datasets for fine-tuning, as our fine-tuning objective is solely directed towards enabling LLAVA to comprehend our task and acquire the ability to describe targets within red detection boxes.

\paragraph{Training Process}
Firstly, it is crucial to emphasize that in this paper, LLAVA does not refer to the model presented in the original LLAVA paper. Instead, it broadly denotes a multimodal paradigm where image information is abstracted into several tokens and inputted into a Large Language Model (LLM) for cross-modal attention with textual tokens. Our LLAVA architecture utilizes Q-Former to extract 32 tokens from the VIT features of the image, and these 32 tokens are subsequently mapped to the same feature space as textual tokens through a linear layer. During fine-tuning, only the final mapping linear layer is trained. This fine-tuning approach enables the model to learn the salience of the target object within the red bounding box in the input image while ensuring that the overall model avoids catastrophic forgetting. We initialize our model with the weights of MiniGPT-4.

\paragraph{Hyperparameter Configuration and Implementation Details}

We employed Vicuna 13B as our Large Language Model (LLM). The batch size and number of epochs were set to 32 and 30, respectively. The initial learning rate was set to $3\times 10^{-5}$. Model training was conducted using 4 threads on a single A100 80GB GPU.

\subsection{Chain-of-Thought and Reflection}

After fine-tuning, the LLAVA model still exhibits a certain degree of hallucinations, manifesting as the generation of imaginary objects. Taking inspiration from the concept of thought chains in large language models, we manually constructed a thought chain specifically tailored for the task of describing target objects. In Table~\ref{tab:table_lmm_COT}, we illustrate the prompts at each step of the thought chain: 1) Inquire about the category of the target object. 2) Inquire about the attributes possessed by the target object. 3) Inquire about the objects surrounding the target object. 4) Inquire about the relationship between the target object and its surrounding objects. Through these four steps, we can attain a comprehensive understanding of LLAVA's perception of the target object. However, hallucinations may still persist in the thought chain; therefore, in step 5), we prompt the model to reexamine the image and correct any errors. Finally, in step 6), we prompt LLAVA to produce the ultimate description of the target object.

\begin{table}[h]
\begin{tcolorbox}[colback=gray!10,
                  colframe=black,
                  width=14cm,
                  arc=3mm, auto outer arc,
                 ]

\begin{itemize}
\item [1)] ``What is the object inside the red bounding box?"

\item [2)]  ``What are the attributes of the object within the red bounding box?"
\item [3)]  ``Which objects are around the target object in the red box?"

\item [4)]  ``What is the relationship between the object inside the red bounding box and the surrounding objects?"

\item [5)]  ``Please review the image once again, and if there are any inaccuracies in your previous answers regarding the object's attributes and relationships, kindly correct them."

\item [6)]  ``Look carefully at the objects in the red box and describe them in one sentence to distinguish them from other objects."

\end{itemize}
\end{tcolorbox}
\caption{Text prompts for guiding chain-of-thought and reflection.}
\label{tab:table_lmm_COT}
\end{table}

\clearpage

\section{Expression Generation via GLOBAL PROMPT PIPELINE}\label{sec:supp-single-modality}

We choose an image from Objects365 as an example to illustrate the instruction generation pipeline via our global prompt pipeline. This pipeline consists of inputs, two steps, in-context samples, and LLaMA outputs, which are illustrated one-by-one in the following:\\

\begin{figure}[h]
    \centering
    \includegraphics[width=\textwidth]{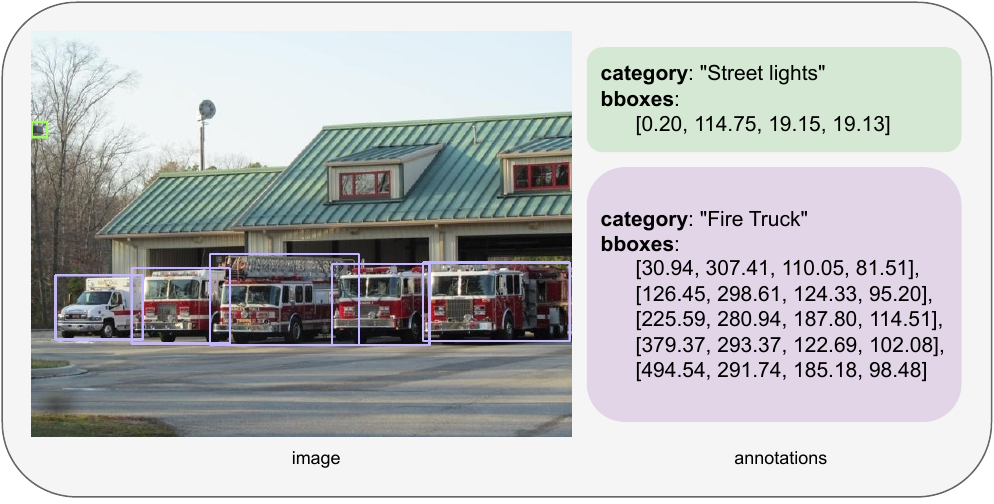}
    \caption{An input example. The left part is an image from the Objects365 dataset. The right part is the object category names and corresponding bounding boxes coordinates.}
    \label{fig:Obj365_input_example}
\end{figure}

\paragraph{Inputs} Our input is an example image with object bbox coordinates, which is shown in Fig.~\ref{fig:Obj365_input_example}.

\paragraph{Output} We send the final text prompt shown in Table~\ref{tab:table_llm_prompt_pseudo} to LLaMA. Table~\ref{tab:obj365_llama_out} shows the output expression of the inputs shown in Fig.~\ref{fig:Obj365_input_example}. The detailed steps to prepare text prompts are illustrated as follows:

\begin{table}[h]
\begin{tcolorbox}[colback=gray!10,
                  colframe=black,
                  width=14cm,
                  arc=3mm, auto outer arc,
                 ]
\textbf{[Fire Truck]}

(1) vehicle, emergency vehicle, fire engine, parked outside the fire station, an essential part of the fire station's resources, essential part of the fire station's resources \\
(2) lined up in a neat row, ready for use, object parked in the row with other fire trucks, object with ladders and equipment 

\textbf{[Street Lights]}

(1) light fixtures, outdoor lighting, two lights visible, providing illumination, source of illumination \\
(2) objects providing illumination, objects in the parking lot, objects providing a clear representation of the overall setting, objects providing light for the parking lot

\end{tcolorbox}
\caption{LLaMA output for the inputs shown in Fig.~\ref{fig:Obj365_input_example}.}
\label{tab:obj365_llama_out}
\end{table}

Figure~\ref{fig:Obj365_input_example} and Table~\ref{tab:obj365_llama_out} depict the original input and final output of the global prompt pipeline, with the intermediary processing steps outlined below. Initially, the generation of the desired caption description for images lacking dense captions is required (\textbf{Step 1}). Subsequently, with the image fully textualized, we employ a detailed prompt design and seed example design to instruct LLAMA in outputting instructions describing the target object (\textbf{Step 2}).

\clearpage

\paragraph{Step 1} 
We use LLaVA to generate image captions. Table~\ref{tab:llava_basic_prompts} shows the text prompt list we use. One random text prompt from this list is combined with object content description (i.e., referring expression or category name) for LLaVA input. For each image, we send these prompts into LLaVA twice and obtain two diverse text descriptions. These two text descriptions are both utilized in the following steps. Examples of generated text descriptions are shown in Table~\ref{tab:obj365_llava_example}.

\begin{table}[h]
\begin{tcolorbox}[colback=gray!10,
                  colframe=black,
                  width=14cm,
                  arc=3mm, auto outer arc,
                 ]
\begin{itemize}
\item   ``Describe the following image in detail" 
\item   ``Provide a detailed description of the given image"
\item    ``Give an elaborate explanation of the image you see"
\item    ``Share a comprehensive rundown of the presented image"
\item    ``Offer a thorough analysis of the image"
\item    ``Explain the various aspects of the image before you",
\item    ``Clarify the contents of the displayed image with great detail"
\item    ``Characterize the image using a well-detailed description"
\item    ``Break down the elements of the image in a detailed manner"
\item    ``Walk through the important details of the image"
\item    ``Portray the image with a rich, descriptive narrative"
\item    ``Narrate the contents of the image with precision"
\item    ``Analyze the image in a comprehensive and detailed manner"
\item    ``Illustrate the image through a descriptive explanation"
\item    ``Examine the image closely and share its details"
\item    ``Write an exhaustive depiction of the given image"
\end{itemize}
\end{tcolorbox}
\caption{Text prompts we use for LLaVA. These prompts follow the author usage in the original LLaVA work.}
\label{tab:llava_basic_prompts}
\end{table}

\begin{table}[h]
\begin{tcolorbox}[colback=gray!10,
                  colframe=black,
                  width=14cm,
                  arc=3mm, auto outer arc,
                 ]
\textbf{Prompt for LLaVA} \\
Provide a detailed description of the given image, including objects: \textit{Street lights}, \textit{Fire Truck} \\

\textbf{Output from LLaVA} \\
\textcolor{blue}{Output 1st} The image displays a parking lot outside a fire station where several fire trucks are parked. In total, there are five fire trucks of varying sizes, all lined up neatly in the lot. Additionally, there are two street lights visible in the scene, providing illumination for the area. The fire trucks appear ready for use and are an bessential part of the fire station's resources. \\
\textcolor{blue}{Output 2nd} The image depicts a parking lot outside a fire station, where several fire trucks are parked in a neat row. There are a total of five fire trucks of varying sizes, all aligned and ready for use. In addition to the fire trucks, there are two street lights visible in the scene, providing illumination for the parking lot. The overall setting gives a clear representation of the organized and prepared nature of the fire station.
\end{tcolorbox}
\caption{Generated examples of LLaVA based on one random text prompt from Table~\ref{tab:llava_basic_prompts}.}
\label{tab:obj365_llava_example}
\end{table}

\paragraph{Step 2} 
We write text prompt for LLaMA. In this step, the text prompt contains three parts including task descriptions, in-context examples, and text description of image (i.e., Table~\ref{tab:obj365_llava_example}). An example text prompt for LLaMA is shown in Table~\ref{tab:table_llm_prompt_pseudo} where an in-context sample is shown in Table~\ref{tab:instruct_seed}.

\clearpage

\begin{table}[h]
\begin{tcolorbox}[colback=gray!10,
                  colframe=black,
                  width=14cm,
                  arc=3mm, auto outer arc,
                 ]
\begin{lstlisting}[language=Python]
task_description = f"""
\end{lstlisting}
\textit{\# Task description prompt}\\

\#\# Establishing fundamental roles and task localization for LLM.

You are an AI visual assistant that can analyze a single image. 

\#\# Elucidating the form and significance of input information.

User will give you several sentences, describing the same image you are observing. In addition, specific interested object locations within the image are given, along with detailed coordinates. These coordinates are in the form of bounding boxes, represented as (x1, y1, x2, y2) with floating numbers ranging from 0 to 1. These values correspond to the left top x, left top y, right bottom x, and right bottom y.

\#\# Explicating the output content and the encapsulated information.

Using the provided caption and bounding box information, give descriptions about the visual content of each interested objects as if you are seeing the image, as an assistant: \\

(1) give descriptions about the object itself, including \textbf{object types, object functions, object counts, object locations, object actions}, etc. \\
(2) give descriptions about the object and other objects, including \textbf{the relative positions between objects, the interaction between objects in the image}, etc.

\#\# Emphasizing common output issues.

Descriptions should be a series of phrases, not whole sentence. Give descriptions for specific interested objects only, do not centered on other objects.\\
Again, give descriptions centered on specific objects only.
\begin{lstlisting}[language=Python]
"""

image2text = f"""
\end{lstlisting}

\textit{\# Image description of LLaVA, the following two paragraphs are from Table~\ref{tab:obj365_llava_example}}\\

The image displays a parking lot outside a fire station where several fire trucks are parked. In total, there are five fire trucks of varying sizes, all lined up neatly in the lot. Additionally, there are two street lights visible in the scene, providing illumination for the area. The fire trucks appear ready for use and are an bessential part of the fire station's resources.\\

The image depicts a parking lot outside a fire station, where several fire trucks are parked in a neat row. There are a total of five fire trucks of varying sizes, all aligned and ready for use. In addition to the fire trucks, there are two street lights visible in the scene, providing illumination for the parking lot. The overall setting gives a clear representation of the organized and prepared nature of the fire station.\\

Street lights: [0.0, 0.23, 0.03, 0.26]\\
Fire truck: [0.05, 0.6, 0.21, 0.76], [0.19, 0.58, 0.37, 0.77], [0.33, 0.55, 0.61, 0.77], [0.56, 0.57, 0.74, 0.77], [0.72, 0.57, 1.0, 0.76]
\begin{lstlisting}[language=Python]
"""
\end{lstlisting}

\textit{\# Python code together with above text prompts are directly sent to LLaMA}
\begin{lstlisting}[language=Python]
messages = [{"role": system, "content": task_description}]
for in_context_example in in_context_examples:
    messages.append({"role": user, "content": in_context_example["content"]})
    messages.append({"role": assistant, "content": in_context_example["response"]}
messages.append({"role": user, "content": image2text})
\end{lstlisting}
\end{tcolorbox}
\caption{An example of prompt generation for LLaMA input. For each image, we first obtain text description in Table~\ref{tab:obj365_llava_example}. Then, we organize the above prompts and python code, together with in-context examples for LLaMA input. An in-context example is in Table~\ref{tab:instruct_seed}.}
\label{tab:table_llm_prompt_pseudo}
\end{table}

\clearpage

\begin{table}[h]
\begin{tcolorbox}[colback=gray!10,
                  colframe=black,
                  width=14cm,
                  arc=3mm, auto outer arc,
                 ]
\begin{wrapfigure}{r}{4.5cm}
    \centering
    \includegraphics[width=0.35\textwidth]{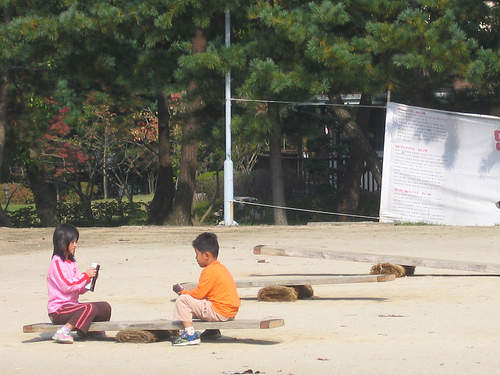}
\end{wrapfigure}

\textbf{\color{brown}\large Example image2text} \\
Two children, a girl and a boy are practicing their writing. \\
Two children sit on a small seesaw in the sand. \\
Two children sitting on a teeter totter. \\
2 kids playing on a seesaw. \\
Two kids sit on a seesaw. \\
2 kids/two kids/two children: [0.09, 0.59, 0.49, 0.94] \\
girl: [0.09, 0.59, 0.23, 0.92] \\
boy: [0.34, 0.62, 0.49, 0.94] \\
seesaw/small seesaw/teeter totter: [0.06, 0.83, 0.57, 0.93] \\
sand: [0.01, 0.61, 1.0, 1.0] \\
\rule{13cm}{0.4pt}
\begin{spacing}{1.5}
\textbf{\color{brown}\large Example Response}
\end{spacing}

\textbf{[2 kids/two kids/two children]} \\
(1) persons, children, two kids, children learning, kids playing, kids sitting, children practicing their writing \\ 
(2) 2 kids playing on a seesaw, Two kids sitting on a seesaw, a girl and a boy, two person on the teeter totter, kids by the sea \\
\textbf{[girl]} \\
(1) girl, kid on the left of the image, girl sitting, girl playing \\
(2) kid playing with the boy, girl sitting on the small seesaw, girl playing in the sand, girl reading a book, a friend of the boy on the right, the taller kid \\
\textbf{[boy]} \\
(1) boy, kid on the right, boy sitting, boy playing, boy practicing his writing \\
(2) kid playing with the girl, boy sitting on a teeter totter, a friend of the girl on the left, boy playing in the sand \\
\textbf{[seesaw/small seesaw/teeter totter]} \\
(1) small seesaw, teeter totter, item to be played on, common facilities in parks and playground, game of two people \\
(2) seesaw in the sand, item the kids are sitting on, item the girl is sitting on, item the boy is playing on \\
\textbf{[sand]} \\
(1) common by the sea, the background of the scene \\
(2) item on which the seesaw is placed, item on which the kids are standing \\
\end{tcolorbox}
\caption{An in-context example. We manually prepare three in-context examples and show one here. On the top block there are image captions and object bbox coordinates. Note that the displayed image is only for image caption reference and is not used in practice. The bottom block shows our expected expressions for each object. The expressions listed in (1) focus on the object attribute itself, and listed in (2) focus on relationship between current object and other objects in this image.}
\label{tab:instruct_seed}
\end{table}

The in-context example in Table~\ref{tab:instruct_seed} follows the task description in Table~\ref{tab:table_llm_prompt_pseudo} and primarily serves to standardize the output format of LLAMA. The desired output format is intended to resemble the \textbf{Example Response} presented in Table 9. [name] denotes the target currently being described, where the descriptions in (1) are relatively simple, pertaining only to the inherent attributes, and those in (2) entail more complex descriptions involving surrounding objects. This facilitates accurate correspondence between each description and the object it pertains to when parsing the output descriptions.

It is noteworthy that the prompt design is manually crafted. It draws significant inspiration from the prompt framework used by LLAVA, with specific modifications and supplements tailored to the requirements of our task. The entire prompt design underwent numerous iterations based on the output results, ensuring the stability of the output format and style.

\clearpage

\section{Visual Prompting and Visual-Textual Matching}\label{sec:supp-vprompt}

\begin{figure}[h]
\begin{center}
    \includegraphics[width=0.9\textwidth]{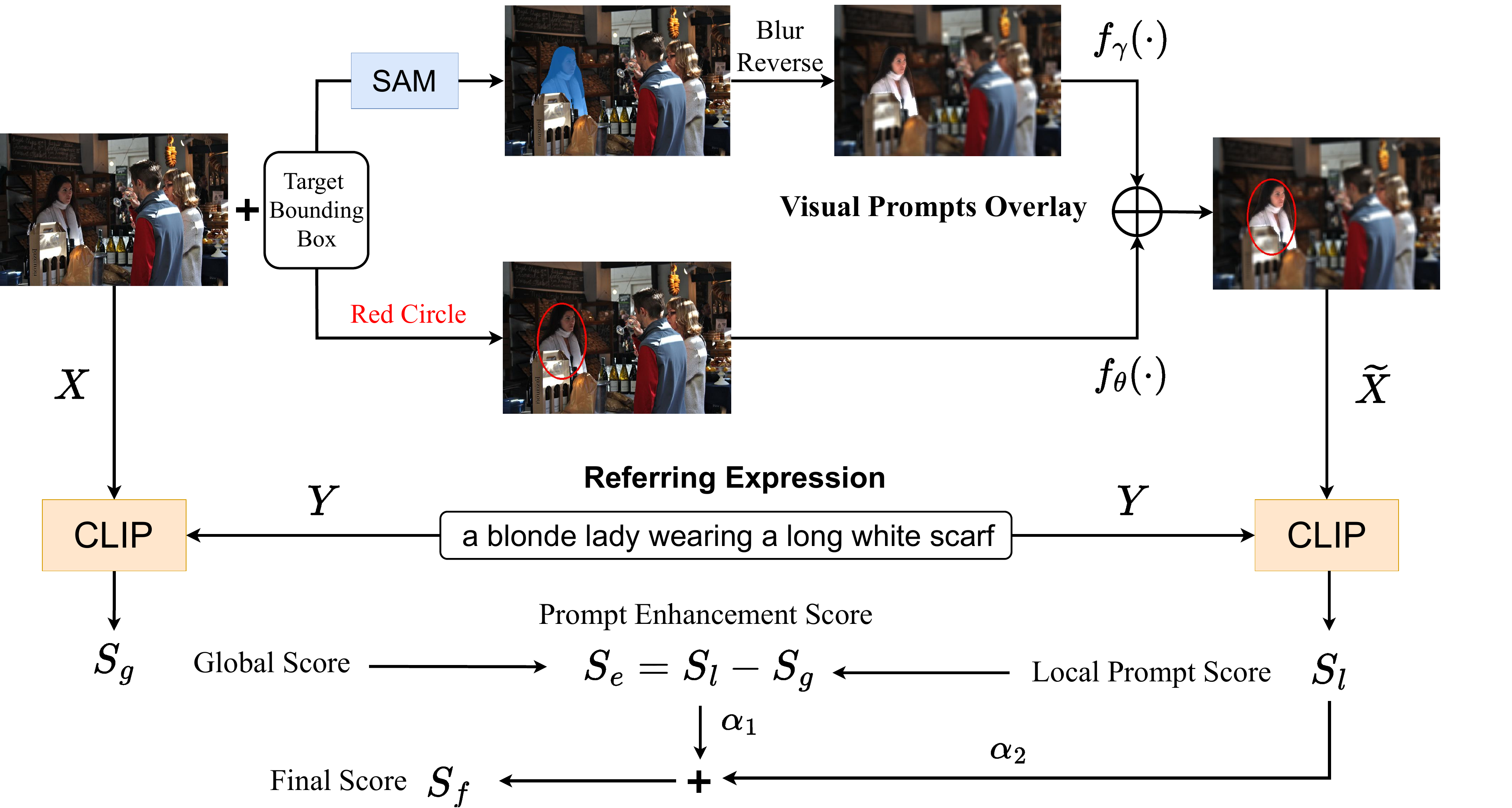}
    \caption{A detailed pipeline of visual prompting and visual-textual matching. We use visual prompting to emphasize target object in the image. Then, we use the CLIP model to compute the scores for expression filtering}.
    \label{fig:clip_filter}
\end{center}
\end{figure}

Motivated by recent applications \citep{FGVP, red_circle} of visual language pretraining (VLP) models in solving zero-shot REC tasks, we discover that coupling VLP with visual prompting enables robust pairing of image regions and text on the basis of potent generalization. The efficacy of pairing is largely influenced by the configuration of visual prompting, which can be divided into two types as follows:

(\romannumeral1) Shape division of emphasized areas: box, circle and contour. 

(\romannumeral2) Highlighting methods for emphasized areas: crop-based, line-based, mask-based, grayscale reversion and blur reverion based image editing operations.

As shown in Fig.~\ref{fig:clip_filter}, after various combinations of experiments, we find the superposition of line-based circle prompt and blur reverse contour prompt yields the optimal visual prompting strategy. The image $\widetilde{X}$ after visual prompting can be expressed as:
\begin{equation}
\widetilde{X} = f_{\theta}(f_{\gamma}(X)) \quad \text{with} \quad f_{\gamma}(\cdot) = BR(\cdot, SAM(\cdot, B)) \quad f_{\theta}(\cdot) = RC(\cdot, B) 
\end{equation}
where $B$ represents the box coordinates of the target, $BR$ and $RC$ represent the reverse Gaussian blur for the mask area and the inscribed ellipse for the target bounding box, respectively. $SAM$ denotes Segment Anything Model \citep{SAM}, which makes visual prompting more refined. 

Once we obtain image $x$ and visually prompted $\widetilde{X}$, we send both of them to the CLIP model together with the generated expressions $Y$ from global prompt and local prompt pipelines. Then, we can compute the global score $S_g$ and local score $S_l$, and thus obtain the final score $S_f$ as illustrated in equation~\ref{eq:dropout} for expression filtering.  

\clearpage

\section{Instruction Generation for Mining Clusters of Multi-objects}\label{sec:supp-multiobj}

In order to find and summarize the common attributes among multiple objects in the clusters from DBSCAN~\citep{DBSCAN}, we use LLaMA for a further analysis. Table~\ref{tab:summary_promts} shows the text prompt of task description and in-context examples we use for the LLaMA inputs. Then LLaMA produces expressions for multi-objects as shown in Fig.~\ref{fig:multi_object}. 

\begin{table}[h]
\begin{tcolorbox}[colback=gray!10,
                  colframe=black,
                  width=14cm,
                  arc=3mm, auto outer arc,
                 ]
\textbf{Task Description}

You are an AI language assistant that can analyze phrases and sentenses.
User will give you descriptions of several objects in an image. Descriptions of each object are several phrases or short sentences.\\

The given objects are expected to have similar properties. Based on the descriptions, find the common properties between given objects and summerize precisely as an assistant: common properties between objects can include same types, same functions, same color components, same poses, same relationships with other objects, engaging in the same activity, etc.\\

If there are no common properties between given objects, just tell that there are no common properties. Your summery should also be phrases. Do not repeat.\\

Give similarity between all given objects, contrary properties like different positions or different colors of clothes should not be included together in your descriptions. \\

\textbf{One In-context Example}

Prompt:
\textcolor{blue}{Objects and their descriptions:\\
\#\# object 2: girl sitting on bed, girl with toy, girl sitting on bed \\
\#\# object 3: man looking down, boy sitting on the bed, man sitting on bed \\
Please find an summarize the similar properties of given objects.}

Response:
\textcolor{blue}{Summary of common properties of given objects: \\
\#\# people on bed; person sitting on bed; people playing on bed; who sitting on bed;}

\end{tcolorbox}
\caption{Task description and an in-context example for multi-objects instruction generation.}
\label{tab:summary_promts}
\end{table}

\clearpage

\section{Instruction Grouping}\label{sec:supp-group}
We use LLaMA to analyze different extents of description based on object category, attribute, and relations. For each instruction of single object in the InDET dataset, we use LLaMA to assign it into one of the preset 4 groups. Table~\ref{tab:leveling_promts} shows the text prompts of task description and in-context examples we use for instruction grouping. For instructions of multiple objects, we assign them to G5 if there is the combination (e.g., ``and'') of single instructions, or we assign them to G6 if the instructions are generated without concatenation.

\begin{table}[h]
\begin{tcolorbox}[colback=gray!10,
                  colframe=black,
                  width=14cm,
                  arc=3mm, auto outer arc,
                 ]
\textbf{Task Description}

You are an AI language assistant that can analyze the language complexity of sentences or phrases.\\
User will give you a phrase or sentence describing a specific object in a image, which could be composed of nouns, adjectives, verbs, etc. \\
Grade the description according to its language complexity as an AI language assistant.\\

The language complexity of a phrase or sentence describing a specific object in an image can be graded into four levels:\\
\textbf{Level 0.} A single noun is used to give the object's name or type.\\
\textbf{Level 1.} A phrase with one or more nouns, verbs and abjectives is used to describe simple attributes of the object itself, such as its color, its function or purpose, location in the image, or actions.\\
\textbf{Level 2} A phrase with one or more nouns, verbs and abjectives is used to describe the object by referring to other objects in the image and describing their relationship, such as their relative positions or interactions.\\
\textbf{Level 3.} A long phrase or sentence is used to describe attributes of the object and also refer to a few other objects in detail, or describe a complicated or comprehensive/implicit relationship between multiple objects.\\
The level of descriptions increase as the language complexity and length increase, and also increase as the phrases or sentences become more descriptive and use more abjectives and nouns to describe the object.\\

\textbf{One In-context Example} \\
Prompt:
\textcolor{blue}{Grade description: people who are sitting under an umbrella.}\\
Response:
\textcolor{blue}{My grading for description people who are sitting under an umbrella: This phrase is referring to the object of people, and gives simple object action of sitting and object relationship with the umbrella. The level of this description is: level 2.}

\end{tcolorbox}
\caption{Task description and in-context examples for single-object instruction grouping.}
\label{tab:leveling_promts}
\end{table}

\clearpage

\section{Model and Implementation Details}
\label{sec:supp-model}

In this section, we illustrate our model architecture and training configurations.

\subsection{Model Architecture}

\begin{figure}[h]
\begin{center}
    \includegraphics[width=0.9\textwidth]{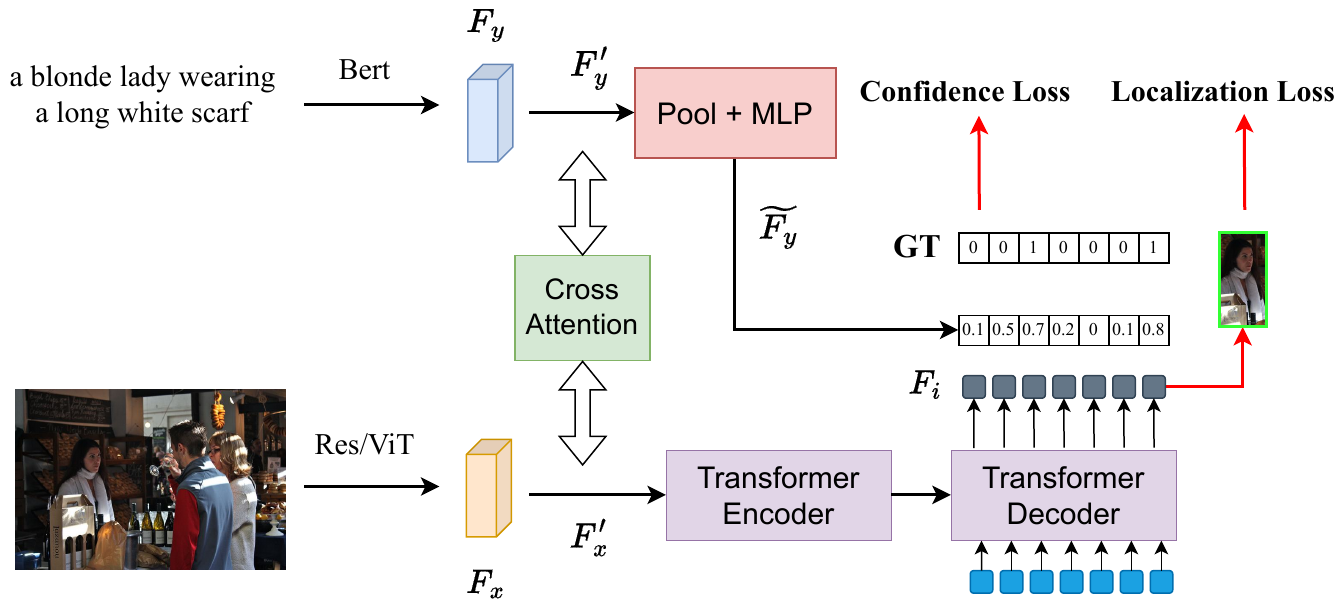}
    \caption{A detailed overview of our diversified referring object detection (DROD) model.}
    \label{fig:baseline_model_detail}
\end{center}
\end{figure}

In the main text, We briefly introduced the DROD model architecture we use. Here, we introduce the overall model structure in Fig.~\ref{fig:baseline_model_detail}. 

\paragraph{Visual Textual Feature Fusion}
We use a general text encoder~\citep{bert} and vision encoder~\citep{VIT} to obtain text features $F_y$ and image features $F_x$. In order to improve the attention of visual contents on the described regions by the text expressions, we conduct a multi-modality feature fusion. Specifically, we use a bidirectional cross attention (Bi-XAtt) module to enhance image and text features through information transfer between modalities. The enhanced image and text features are then added to the original features. This process can be formulated as follows:
\begin{equation}
F_x' = F_x + F_{y2x}\; ; \; F_y' = F_y + F_{x2y} \;\;\;\;  \text{with} \;\;\;\;  F_{y2x}, F_{x2y} = \text{Bi-XAtt}(F_x, F_y)
\end{equation}

\paragraph{Target Discovery and Retrieval}
With the enhanced visual and language representations, we need to extract the targets referred to by the expression from the image features. Our DROD model applies the encoder-decoder architecture of Deformable DETR~\citep{deformable}, which allows for more flexible query retrieving. Here we provide a detailed introduction to our internal module. 

The transformer encoder receives multi-scale visual features after text enhancement. Multi-scale Deformable encoder utilizes flexible attention at different scales and spatial shapes to obtain hierarchical characteristics for instances. In addition, following the design of two-stage Deformable DETR, we add an auxiliary prediction head for reference points. The top $N$ points are input into the decoder as the position prior.

The Transformer decoder uses learnable queries to retrieve instance-related information from encoded multi-scale visual features. The design of the query is critical for ensuring stability and efficiency in training, based on previous works \citep{trackformer,VIST}. The $N$ reference points served by encoder act as the position priors of the $N$ queries. The content part of each query is a static learnable vector. Moreover, following DINO, we add denoising queries to make the decoder's convergence more stable and faster. Through Deformable attention, $N$ queries efficiently retrieve instance embedding $F_{i} \in \mathbb{R}^{N \times d}$ from expression-aware visual features. 

Finally, we need to select the instances referred to by expression from the $N$ instance proposals. As shown in Fig.~\ref{fig:baseline_model_detail}, we use global average pooling and MLP to map the visual-aware expression feature $F_y'$ to the semantic space of the query embedding to obtain $\widetilde{F_y} \in \mathbb{R}^{1 \times d}$. Thus, the matching score between each query proposal and the expression can be expressed by the cosine similarity between vectors. $S = F_{i} \times \widetilde{F_y}^\top$. In the inference stage, proposals with scores above threshold $\theta$ are selected. This flexible selecting strategy allows our architecture to output any number of results that satisfy user requirements. In the training stage, the overall model is supervised by a combination of confidence loss and localization loss.

\subsection{Training Configurations}

\paragraph{Generation Engine}
$\alpha_1$ in the expression filter responsible for balancing referentiality and semantic accuracy is set to 0.5. While generating multi-target expressions, DBSCAN clustering method has two hyperparameters: neighbourhood radius $eps$ is 1.5 and minimum neighbors of the core point $minPts$ is 2. The temperatures of LLaMA are set to 0.7.

\paragraph{DROD Model}
We use ResNet-50~\citep{resnet} and ViT-Huge~\citep{VIT} as visual encoders and Bert \citep{bert} as the text encoder. The transformer encoder-decoder architecture consists of a six-layer encoder and a six-layer decoder. The number of object queries $N$ is set to 900. Our DROD model is initialized by weights pretrained on Objects365 released by UNINEXT~\citep{UNINEXT}. The optimizer we use is AdamW \cite{adamw} with a learning rate of 2e-4, a weight decay of 0.05 and the warm-up steps are 400 with an initial learning rate of 4e-5. The model is trained on 32 and 16 V100 GPUs for pretraining and finetuning, respectively.

\clearpage

\section{Ablation Studies}
\label{sec:supp-ablation}
In this section, we study the necessity of finetuning LLaVA during expression generation from  local prompt pipeline. Then, we investigate how different visual prompting strategies affect the CLIP model to maintain our expected expressions.

\paragraph{LLaVA Finetuning}
In our InstructDET, we do not finetune LLaMA in global prompt pipeline, but finetune LLaVA in local prompt pipeline. There are two main reasons illustrated as follows:

\begin{figure}[h]
    \centering
    \includegraphics[width=0.8\textwidth]{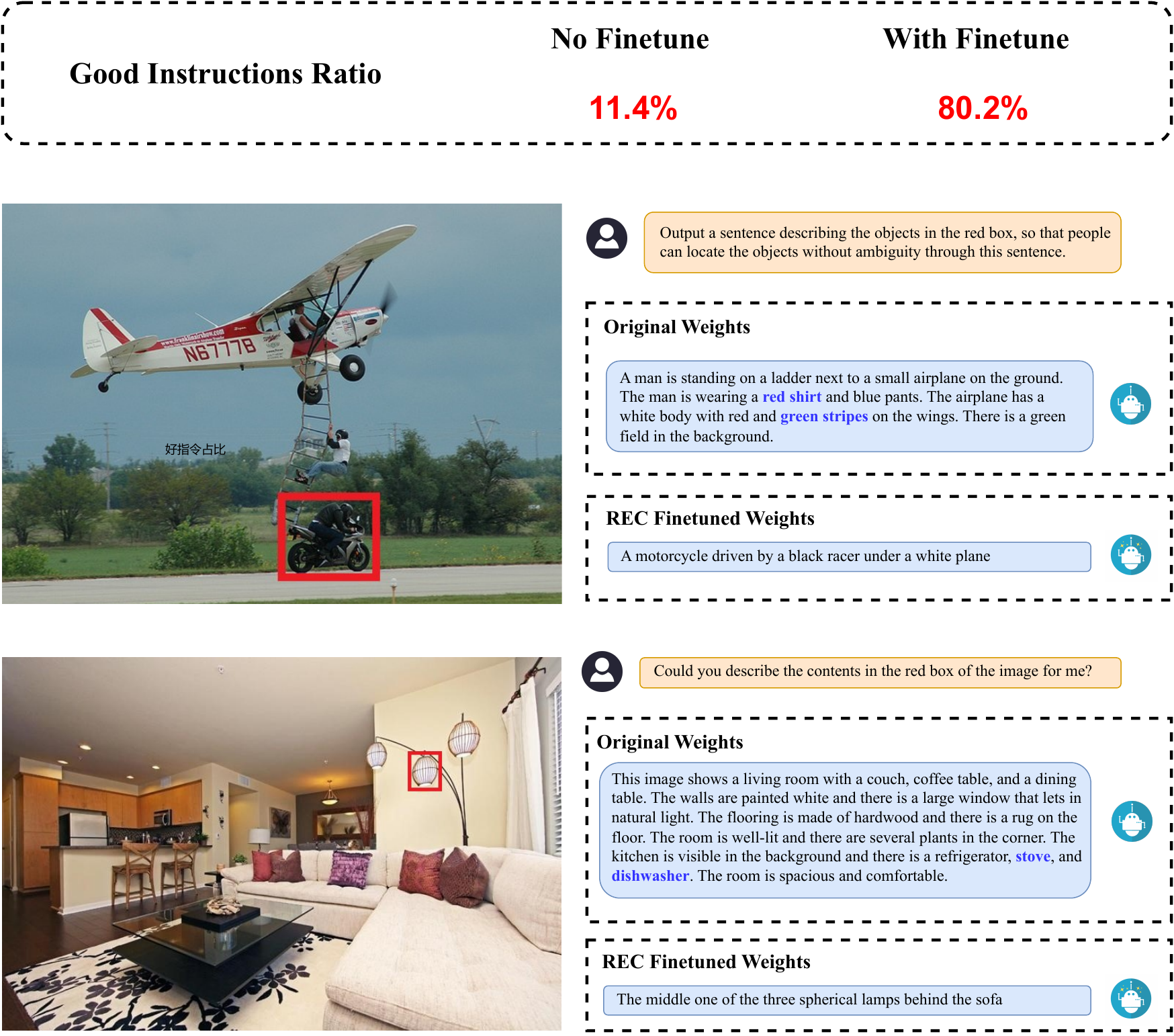}
    \caption{Comparison of expressions generated by LLaVA before and after our finetuning. Blue words indicate incorrect descriptions.}
    \label{fig:finetune_LMM}
\end{figure}

(\romannumeral1) Asking questions directly to LLaVA will get the answers with dense description, rather than the exact expressions we expect for the specific target objects.

(\romannumeral2) Due to the limited number of visual tokens output by Q-Former, it is difficult to fully display all the visual features of the entire image. With limited visual information, LLaVA produces lengthy descriptions and thus lead to massive hallucinations (blue words in Fig.~\ref{fig:finetune_LMM}). The generated expressions in this way becomes unstable.

Based on above analysis, we partially finetune LLaVA on the linear layer that that transforms visual features into the semantic text embedding space. To this end, the model is updated to understand which area we aim to emphasize (i.e., marked in the red box in Fig.~\ref{fig:finetune_LMM}) and which visual features are important for target object description. Fig.~\ref{fig:finetune_LMM} shows that after finetuning the linear layer using REC datasets, the probabilities of generating our expected high-quality expressions increases from 11.4\% to 80.2\%.

\begin{table}[h]
\small
\centering
\caption{Visual Prompting (VP) Evaluation. Different VP methods make differences in the ability of CLIP to retrieve high-quality expressions.}
\label{tab:visual prompting}
\begin{tabular}{c|cc|cc}
\hline
\multirow{2}{*}{Strategy} & \multicolumn{2}{c|}{Visual Prompting} & \multicolumn{2}{c}{Expression Retrieval (\%)} \\
 & Shape & Tool & Easy $\uparrow$ & Hard $\uparrow$ \\ \hline
1 & \multirow{5}{*}{Box} & Crop & 31.52 & 24.09 \\
2 &  & Gray & 39.66 & 31.85 \\
3 &  & Line & 50.12 & 42.30 \\
4 &  & Mask & 48.29 & 39.93 \\
5 &  & Blur & 52.75 & 44.30 \\ \hline
6 & \multirow{3}{*}{Circle} & Line & 52.78 & 44.33 \\
7 &  & Mask & 51.01 & 42.87 \\
8 &  & Blur & 54.13 & 46.22 \\ \hline
9 & \multirow{3}{*}{Contour} & Line & 52.34 & 44.05 \\
10 &  & Mask & 51.90 & 43.44 \\
11 &  & Blur & 55.79 & 47.78 \\ \hline
12 & \multicolumn{2}{c|}{VP3 + VP11} & 56.81 & 48.97 \\
13 & \multicolumn{2}{c|}{\textbf{VP6 + VP11}} & \textbf{58.29} & \textbf{50.99} \\ \hline
\end{tabular}
\end{table}

\paragraph{Visual Prompting Selection}
The selection of visual prompting (VP) is crucial for expression filtering. We evaluate the affect of various visual prompting methods based on the retrieval performance of CLIP model in Table~\ref{tab:visual prompting}. First, we define a new metric called expression retrieval ratio, which indicates the proportion of correct expressions that can be retrieved based on the visual-textual matching of CLIP model. During testing, each minibatch contains $k$ correct target expressions, and the remaining expressions are negative samples. We take the expressions with the top $k$ Clip score as the retrieved expressions. Finally, the proportion of correct expressions in the retrieved expressions is the expression retrieval ratio. Under the Easy setting, the negative samples in the minibatch come from other images. Under the Hard setting, the negative samples in the minibatch may come from different targets in the same image. Table~\ref{tab:visual prompting} shows that cropping (1) and grayscale reversion (2) methods achieve poor results. Because cropping loses all surrounding environment information, and grayscale reversion loses all color features. The best single prompt is the contour blur reversion. The best combination prompt is the contour blur reversion and circle line. Circle line can indicate the rough areas that needs attention, and contour blur reversion can highlight the target object in a fine-grained manner that eliminates background interference. Inevitably, the prior knowledge in the CLIP model is also important to the difference in visual prompt effects. In CLIP's pre-trained web-scale dataset, images with red circles often indicate that the targets in the red circles are more important and need to be noticed. A large amount of photography also exists in the web-scale dataset, which employs “Bokeh” to blur the background and highlight the subject.

\clearpage

\section{SYNONYMOUS REWRITE}
\label{sec:supp-syn-rewrite}
In post processing, we utilize LLaMA to do synonymous rewriting to further diversify the generated expressions. The prompt we used is shown in Table~\ref{tab:syn_rewrite_promts}.

\begin{table}[h]
\begin{tcolorbox}[colback=gray!10,
                  colframe=black,
                  width=14cm,
                  arc=3mm, auto outer arc,
                 ]
\textbf{Synonymous Rewriting Prompt}

I want you to act as a synonymous expression provider. I will give you a text of phrase or short sentence, which is an expression that describes a main object while mentioning some other objects. And you will reply to me with a new expression that have the same semantic meaning and describe the same main object as the provided expression. The new expressions should also be phrases or short sentences no longer than 25 words. Do not write explanations on replies. Do not repeat.

\end{tcolorbox}
\caption{Synonymous Rewriting Prompt for LLaMA.}
\label{tab:syn_rewrite_promts}
\end{table}

\section{Supplement Comparisons}
\label{sec:supp-comparison}

\subsection{Evaluation on InDET}
\begin{table}[h]
\scriptsize
\centering
\renewcommand{\tabcolsep}{2mm}
\renewcommand\arraystretch{1.2}
\setlength{\belowcaptionskip}{3pt}
\caption{Evaluation results on our InDET. We show the object bbox average precision (AP) values ($\%$).}
\label{tab:eval_on_InDET_appendix}
\vspace{-3mm}
\begin{threeparttable}
\begin{tabular}{l|c|c|cccccc}
\hline
\multirow{2}{*}{Method} & \multirow{2}{*}{Backbone} & \multirow{2}{*}{AP} & \multicolumn{6}{c}{AP by Group}\\
 & & & G1 & G2 & G3 & G4 & G5 & G6 \\ \hline
 MDETR & ResNet101 & 34.86 & 47.44 & 46.79 & 34.14 & 23.22 & 25.91 & 28.17 \\
 G-DINO & SwinB & 35.96 & 47.10 & 47.17 & 35.29 & 26.84 & 27.95 & 27.61 \\
 UNINEXT & ResNet50 & 43.37 & 54.49 & 54.52 & 44.49 & 37.17 & 31.41 & 32.01 \\ \hline
 DROD (Ours)\tnote{1} & ResNet50 & 62.24 & 67.14 & 67.34 & 60.89 & 55.10 & 70.15 & 74.22 \\
 DROD (Ours)\tnote{2} & ResNet50 & 62.34 & 67.22 & 68.04 & 61.09 & 55.42 & 68.60 & 72.91 \\
 DROD (Ours) & ViT-H & 66.90 & 72.53 & 72.47 & 66.42 & 59.86 & 73.34 & 75.92 \\ \hline
\end{tabular}
\begin{tablenotes}
\item[1] For fair comparison, our DROD model in Table~\ref{tab:eval_on_InDET} only utilizes RefCOCO/+/g datasets but with diversified instructions, which is partial of InDET. \item [2] Here we add evaluation results of DROD model trained on full InDET.
\end{tablenotes}
\end{threeparttable}
\end{table}

\subsection{Evaluation on RefCOCO/g/+}

\begin{table}[h]
\scriptsize
\centering
\renewcommand\arraystretch{1.2}
\setlength{\belowcaptionskip}{3pt}
\caption{Supplementary Evaluation results on the RefCOCO/g/+ datasets. We follow evaluation protocols to report AP values ($\%$) of comparing methods. We use the notations "CC", "VG", "OI", "O365", "RIGame", for COCO, Visual Genome, OpenImage, Objects365, ReferItGame, respectively.}
\label{tab:eval_on_refcoco_appendix}
\vspace{-3mm}
\begin{threeparttable}
\begin{tabular}{l|c|c|ccc|ccc|cc}
\hline
\multirow{2}{*}{Method} & \multirow{2}{*}{Backbone} & \multirow{2}{*}{Data} & \multicolumn{3}{c|}{RefCOCO} & \multicolumn{3}{c|}{RefCOCO+} & \multicolumn{2}{c}{RefCOCOg}\\
 & & & val & testA & testB & val & testA & testB & val-u & test-u\\ \hline
 RefTR & ResNet101 & VG & 85.65 & 88.73 & 81.16 & 77.55 & 82.26 & 68.99 & 79.25 & 80.01 \\
 SeqTR & DarkNet53 & VG,RIGame,Flickr,RefC & 87.00 & 90.15 & 83.59 & 78.69 & 84.51 & 71.87 & 82.69 & 83.37 \\
 MDETR & ResNet101 & GoldG,CC,RefC & 86.75 & 89.58 & 81.41 & 79.52 & 84.09 & 70.62 & 81.64 & 80.89 \\
 G-DINO & SwinB & O365,CC,RefC,GoldG,etc & 83.95 & 87.79 & 79.16 & 72.91 & 80.91 & 62.96 & 76.98 & 76.76 \\
 PolyFormer & SwinL & GoldG,ReferIt,RefC & 90.38 & 92.89 & 87.16 & 84.98 & 89.77 & 77.97 & 85.83 & 85.91 \\
 UNINEXT & ResNet50 & O365,CC,RefC & 87.64 & 90.35 & 83.49 & 78.14 & 83.22 & 68.71 & 80.96 & 81.86 \\\hline
 DROD (Ours)\tnote{1} & ResNet50 & O365,CC,InDET & 88.92 & 90.86 & 85.57 & 78.27 & 83.39 & 71.04 & 83.01 & 82.91 \\
 DROD (Ours)\tnote{2} & ResNet50 & O365,CC,InDET & 89.85 & 92.03 & 87.24 & 80.50 & 85.87 & 73.61 & 83.93 & 84.73 \\
 DROD (Ours)\tnote{3} & ViT-H & O365,CC,InDET & 92.93 & 94.47 & 91.13 & 86.20 & 89.82 & 80.86 & 88.62 & 89.46 \\\hline
\end{tabular}
\begin{tablenotes}
\item[1] For fair comparison, our DROD model in Table~\ref{tab:eval_on_refcoco} only utilizes RefCOCO/+/g datasets but with diversified instructions, which is a part of InDET. \item [2] Here we add evaluation results of DROD model trained on full InDET. \item [3] DROD here with ViT-H as backbone also utilizes RefCOCO/+/g datasets with diversified instructions.
\end{tablenotes}
\end{threeparttable}
\end{table}

We have supplemented the comparisons with more specialized models, such as Polyformer\citep{polyformer}. With comparable magnitudes in the backbone network, our model continues to exhibit advantages.

\clearpage
\section{Time Consumption}
\label{sec:supp-time-consumption}

\begin{table}[h]
\scriptsize
\centering
\renewcommand\arraystretch{1.2}
\setlength{\belowcaptionskip}{3pt}
\caption{Time Consumption Statistics. We use "instr." as the short for "instruction".}
\label{tab:supp-time-consumption}
\vspace{-3mm}
\begin{threeparttable}
\begin{tabular}{c|c|c|c|c|c}
\hline
 Pipeline & Step & Main Model & BatchSize & Time & Avg \\\hline
 \multirow{2}{*}{Global Prompt} & Image Caption & LLaVA & 4 & 10.71 s/batch & 2.68 s/image \\
  & Instr. Generation & LLaMA & 1 & 17.25 s/image & 0.63 s/instr. \\\hline
 Local Prompt & Instr. Generation & LLaVA & 16 & 2.47 s/batch & 0.15 s/instr.  \\\hline
 \multirow{3}{*}{Post-Processing} & Instr. filtration & CLIP & 1 & 0.186 s/image & 0.016 s/box \\
  & Multi-Object Instr. & DBSCAN & 1 & 0.053 s/image & - \\
  & Instr. Grouping & LLaMA & 1 & 1.43e-06 s/instr. & - \\\hline
\end{tabular}
\end{threeparttable}
\end{table}

 Table~\ref{tab:supp-time-consumption} shows the time consumption of the main steps in our instruction generation procedure. The data is obtained from NVIDIA-v100-32G machines. Obviously, the most time consuming steps are those who depend on large models. Except for the limited hardware performance, another reason why the instruction generation step in global prompt pipeline is very time consuming is that we have very long LLaMA prompts which include three in-context examples and we also desire long response which could include more diversified instruction for each object in image. So this step requires multiple machines to generate simultaneously. While the text received and generated in each forward pass is relatively short in the local prompt pipeline, the generation efficiency is significantly constrained by the fact that only one instruction can be generated per forward pass. Therefore, we address this limitation by maximizing parallelism through the increase in batch size, allowing the simultaneous generation of multiple instructions.

\end{document}